\documentclass[journal]{vgtc}                 

\onlineid{0}

\vgtccategory{Research}

\vgtcpapertype{please specify}

\title{\textit{EvolvED}: Evolutionary Embeddings to Understand the Generation Process of Diffusion Models}

\author{%
  Vidya Prasad, 
  Hans van Gorp,
  Christina Humer,
  Ruud J. G. van Sloun,
  Anna Vilanova, and
  Nicola Pezzotti 
}

\authorfooter{
  \item
  	V. Prasad, H. van Gorp, R. J. G. van Sloun, A. Vilanova, and N. Pezzotti are with Eindhoven University of Technology.
  	E-mail: v.prasad@tue.nl
  \item
  	C. Humer is with Johannes Kepler Universität Linz.
  \item H. van Gorp and N. Pezzotti are with Philips.
}

\abstract{
\remove{Diffusion models generate high-quality samples by corrupting data with Gaussian noise and iteratively reconstructing it with deep learning, progressively transforming noisy images into refined outputs.
Understanding this data evolution is important for interpretability but is complex due to its high-dimensional evolutionary nature. 
Existing dimensionality reduction methods like t-distributed stochastic neighborhood embedding (t-SNE) aid in understanding high-dimensional spaces but neglect evolutionary structure preservation.
Hence, we propose \textit{Tree of Diffusion Life} ($TDL$), a method to understand the evolution of data in the generative process of diffusion models. 
$TDL$ samples a diffusion model's generative space via varied prompts and employs image encoders to extract semantic meaning from these samples, projecting them to an intermediate space. It employs a novel evolutionary embedding algorithm that explicitly encodes the iterations while preserving the high-dimensional relations, facilitating the visualization of the data evolution.
This embedding leverages three metrics: a \textit{standard t-SNE loss} to group semantically similar elements, a \textit{displacement loss} to group elements from the same iteration step, and an \textit{instance alignment} loss to align elements of the same instance across iterations. 
We present rectilinear and radial layouts to represent iterations, enabling comprehensive exploration.
We assess various feature extractors and highlight $TDL$'s potential with prominent diffusion models like GLIDE and Stable Diffusion with different prompt sets. 
$TDL$ simplifies understanding data evolution within diffusion models, offering valuable insights into their functioning.}
\addition{Diffusion models, widely used in image generation, rely on iterative refinement to generate images from noise. 
Understanding this data evolution is important for model development and interpretability, yet challenging due to its high-dimensional, iterative nature. 
Prior works often focus on static or instance-level analyses, missing the iterative and holistic aspects  of the generative path. 
While dimensionality reduction can visualize image evolution for few instances, it does not preserve the iterative structure.
To address these gaps, we introduce \textit{EvolvED}, a method that presents a holistic view of the iterative generative process in diffusion models. 
\textit{EvolvED} goes beyond instance exploration by leveraging predefined research questions to streamline generative space exploration. 
Tailored prompts aligned with these questions are used to extract intermediate images, preserving iterative context. Relevant feature extractors are used to trace the evolution of key image features, addressing the complexity of high-dimensional outputs.
Central to \textit{EvolvED} is a novel evolutionary embedding algorithm that encodes iterative steps while maintaining semantic relations. 
It enhances the visualization of data evolution by clustering semantically similar elements within each iteration with t-SNE, grouping elements by iteration, and aligning an instance's elements across iterations.
We present rectilinear and radial layouts to represent iterations and support exploration.
We apply \textit{EvolvED} to diffusion models like GLIDE and Stable Diffusion, demonstrating its ability to provide valuable insights into the generative process.
}
}


\teaser{
  \centering
  \includegraphics[width=0.99\linewidth]{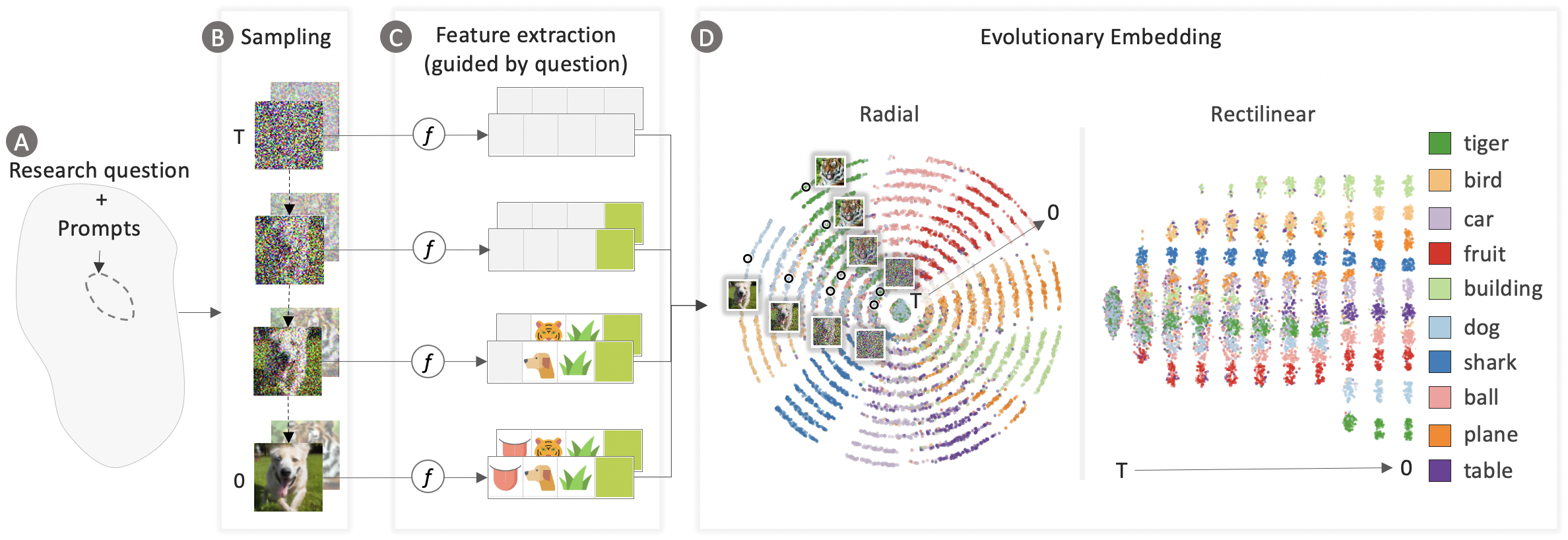}
  \caption{
    \remove{Illustration of the \textit{Tree of Diffusion Life} ($TDL$) method for understanding data evolution in diffusion models. 
$TDL$ samples the generative space of GLIDE~\mbox{\cite{glide}} using $n=1000$ instances across $10$ ImageNet classes as prompts. 
Noisy instance images ($\textbf{h}^i_{t-1}$) are extracted every $10^{th}$ step ($t$ in $[0,100]$) and are projected to an intermediate space ($\widehat{\textbf{h}}{}^i_{t-1}$) via an ImageNet classifier $f$~\mbox{\cite{bai2021transformers}} for semantically meaningful distances. 
Noisy samples $\widehat{\textbf{h}}{}^i_{t-1}$ are visualized using a novel evolutionary embedding method. 
The inner radial ring in the radial layout (left) and the left vertical line in the rectilinear one (right) denote the initial generation steps, progressing outward or towards the right to the final steps. 
Clusters are initially unclear and gradually emerge toward the end; for example, $``$planes$"$ are initially confused with $``$birds$"$ or $``$cars$"$ based on whether they are flying or on the ground, becoming evident only towards the last few steps.}
\addition{Understanding the evolutionary generation process in diffusion models with \textit{EvolvED}. 
(A) Predefined research questions with tailored prompts. 
(B) Example of sampling the generative space of GLIDE~\cite{glide}. 
(C) Semantic object and context features of noisy images ($\textbf{h}^i_{t-1}$) extracted at every $10^{th}$ step using an ImageNet classifier~\cite{bai2021transformers}, capturing key features aligned with the research objectives.
(D) The evolutionary embedding method explicitly encodes iterations. The inner ring in the radial layout and the leftmost line in the rectilinear layout represent the initial generation steps, progressing outward or towards the right to the final steps.
Clusters such as $``$dogs$"$ are initially confused with $``$tigers$"$ due to their presence on grass but gradually become distinct towards the later steps.}
  }
  \label{fig:teaser}
}

\graphicspath{{figs/}{figures/}{pictures/}{images/}{./}} 
\usepackage{soul}   
\usepackage{color}  

\usepackage{tabu}                      
\usepackage{booktabs}                  
\usepackage{lipsum}                    
\usepackage{mwe}                       
\usepackage{footmisc}

\usepackage{mathptmx}                  
\usepackage{algorithm}
\usepackage{algorithmic}
\usepackage{mathrsfs}
\usepackage{amsmath}

\newcommand{\remove}[1]{}
\newcommand\addition[1]{{\color{black}#1}}

\newcommand{\removenew}[1]{}
\newcommand\additionnew[1]{{\color{black}#1}}

\begin{document}

\firstsection{Introduction}
\maketitle 
Diffusion models have gained popularity across various applications~\cite{yang2023diffusionsurvey}, like generation, inverse problems, segmentation, and anomaly detection. 
They effectively generate diverse and high-quality samples by progressively corrupting data with noise, followed by an iterative reconstruction. 
At each step, a trainable DL model (such as U-Net~\cite{unet}) \remove{takes}\addition{processes} an intermediate noisy image, the iteration number, and optionally a prompt\remove{ as input, generating}\addition{, generating} a less noisy image. 
This evolution goes from an initial state of pure noise to a coherent image (see \cref{fig:diffusion_pdf}). 

Analyzing data evolution offers insights beyond image changes, revealing model learnings, data distribution evolution\remove{Understanding this data evolution is relevant to understanding distribution evolution and what the model has learned}~\cite{park2024riemannian}\addition{, and the decision-making dynamics of dataset modes}.\remove{It sheds light on the feature evolution and the decision-making dynamics of dataset modes.}
Such knowledge is valuable for refining models and training~\cite{karras2022elucidating, yang2023disdiff, deja2022analyzing, meng2023distillation}, controlling generation~\cite{park2024riemannian, wu2023uncovering, duan2023schedulerpath}, \remove{and }improving performance~\cite{si2023freeu, prasad2023unraveling}\addition{, and identifying feature entanglements~\cite{yang2023disdiff}.}\remove{. 
This analysis could also help identify the biases and attribute entanglements relevant for robust model development~\mbox{\cite{yang2023disdiff}}.
However, this evolutionary process's iterative and high-dimensional nature presents challenges in comprehending its underlying dynamics.}
\addition{Despite these needs, there is limited support for holistic analysis of the evolutionary process. 

Current explainable AI methods mainly analyze classification models~\cite{prasad2022transform}, which have final-layer 1D representations and single-step outputs. 
In contrast, diffusion models involve iterative evolution and high-dimensional outputs, limiting the value of static analysis methods. 
Additionally, these image-to-image translation models internally lack a single low-dimensional latent representation, complicating the identification of a meaningful feature space for analysis~\cite{prasad2022transform}.
While methods to analyze diffusion models exist, they are limited in their ability to go beyond analyzing the iterative generative process of a single instance.
They often focus on high-level insights, such as feature granularity~\cite{prasad2023unraveling, park2024riemannian}, or explore randomly selected instances~\cite{park2024timeconcepts}, insufficient for a comprehensive understanding. 
Some methods use synthetic low-dimensional datasets~\cite{karras2022elucidating} and local latent structure analyses~\cite{park2024riemannian} to illustrate this process but they offer limited insights into the feature evolution on real data sets.
Dimensionality reduction (DR) techniques~\cite{van2008tsne, umap} commonly used to visualize high-dimensional spaces, can be directly used to show feature evolution over a few instances~\cite{lee2023diffexplainer}.
However, they are not designed to support the analysis of the evolutionary aspect of diffusion data. 
The stochastic cluster placing in DR methods such as t-SNE~\cite{van2008tsne} or UMAP~\cite{umap}, further complicates tracing data evolution and hinders analysis.
There is a need for a method that preserves the iterative structure while providing a holistic view of data evolution to effectively understand and develop diffusion models, as existing methods fail to address both aspects together.}

\remove{Prior works have explored understanding diffusion models. Attribution maps explored concept prioritization of an instance over iterations~\mbox{\cite{park2024timeconcepts}}. 
Methods to understand the generation process within diffusion models, including comparisons of a few instance evolutions, have also been developed~\mbox{\cite{lee2023diffexplainer}}.
While other works delve deeper into the underlying image space in the generative process~\mbox{\cite{park2024riemannian}}, the focus has been to extract a local latent structure around a sample. 
Insights focused on the granularity of features rather than the actual features evolving, which is our goal.
 
Dimensionality reduction (DR) techniques like t-distributed stochastic neighborhood embedding (t-SNE) or uniform manifold approximation and projection (UMAP)~\mbox{\cite{umap}} can visualize feature evolution over few instances~\mbox{\cite{lee2023diffexplainer}}. 
 However, they struggle to scale to large datasets and fail to preserve the iterative structure essential for understanding the diffusion process extending them to visualize the evolution of the entire datasets presents challenges. 
They fail to preserve the iterative diffusion structure crucial for its understanding in \mbox{\cref{fig:diffusion_pdf}}. Further, the stochastic positioning of clusters in these methods complicates tracing this data evolution, hindering analysis.
Preserving the iterative structure is essential for studying data evolution and branching patterns across the entire dataset.}
\addition{To address these gaps, we present \textit{EvolvED}, a method designed to provide a holistic understanding of data evolution within the generative process of diffusion models (see \cref{fig:teaser}). Unlike prior works, \textit{EvolvED} enables a targeted approach relying on user-defined research questions. 
This facilitates a focused investigation into specific characteristics, such as the progression of natural objects or artistic styles.
While the broader aim is to analyze the entire generative space, \textit{EvolvED} narrows its focus to manage the complexity of the space. 
We sample the diffusion model using prompts tailored to the research question.
We extract intermediate images at multiple stages of the generative process to capture the stepwise progression of the model output. 
Since diffusion models internally lack a unified low-dimensional latent space per step, we use external feature extractors to isolate and track attributes of interest aligned with the research question from interim images. 
Finally, we propose a novel \textit{evolutionary embedding} algorithm that enhances t-SNE to be able to capture the evolving structure of generative processes. 
\textit{Evolutionary} refers to the method’s ability to consider the dynamic aspects of the generated data, allowing to trace changes through the iterations.
Our algorithm augments the t-SNE cost function to contain three key components: 1) clustering semantically similar elements, 2) grouping elements of the same iteration, and 3) aligning the same instance across iterations. 
We introduce rectilinear and radial layouts to explicitly represent the iterations.}
\remove{To address the issue of holistic understanding of the generative space, $TDL$ samples this space via several instances with different prompts.
Semantic meaning is extracted from this sampled evolutionary data using feature extractors or image encoders, projecting them to an intermediate space. 
We propose a novel evolutionary embedding algorithm to understand all sampled data at scale while preserving the iterative context, facilitating data evolution visualization. The proposed embedding enables this via three metrics: 1) clustering semantically similar elements, 2) placing elements of the same iteration together, and 3) aligning elements of an instance across iterations. 
Rectilinear and radial layouts explicitly representing these iterations are introduced, allowing a comprehensive exploration of data evolution.
We investigate the strengths and limitations of image encoders that extract semantic meaning from the sampled evolutionary data within the proposed method.}
We demonstrate \addition{$EvolvED$}\remove{$TDL$}'s utility and versatility with two prominent diffusion models, GLIDE~\cite{glide} and StableDiffusion~\cite{stablediff}, \remove{utilizing different prompt set for each model}\addition{across different research questions}. 
By exploring how data and their specific attributes evolve, we gain valuable insights into the generation process of diffusion models. 
Our contributions are:

\begin{itemize}
    \item 
    \remove{$TDL$}\addition{$EvolvED$}, a method to holistically understand data evolution within diffusion models. 
    \remove{It includes sampling the generative space, semantically encoding these samples, and a novel means to visualize the semantic evolution of these samples. }\addition{It includes a pipeline to sample the generative space, extract relevant features, and visualize their evolution during generation}\footnote{The code is available at \url{https://github.com/vidyaprsd/EvolvED}}.
    \item A novel \textit{evolutionary embedding} \addition{algorithm that visualizes the evolutionary nature of diffusion models.}\remove{illustrating the evolution of instances}
    \item \remove{$TDL$}\addition{$EvolvED$}'s 
    \addition{applicability is evaluated via three cases with diverse research questions and models}\remove{evaluation via three cases with diverse prompt sets and diffusion models, to explore its applicability in different scenarios}.
\end{itemize}

\section{Background}
\begin{figure}
    \centering
    \includegraphics[width=0.95\linewidth]{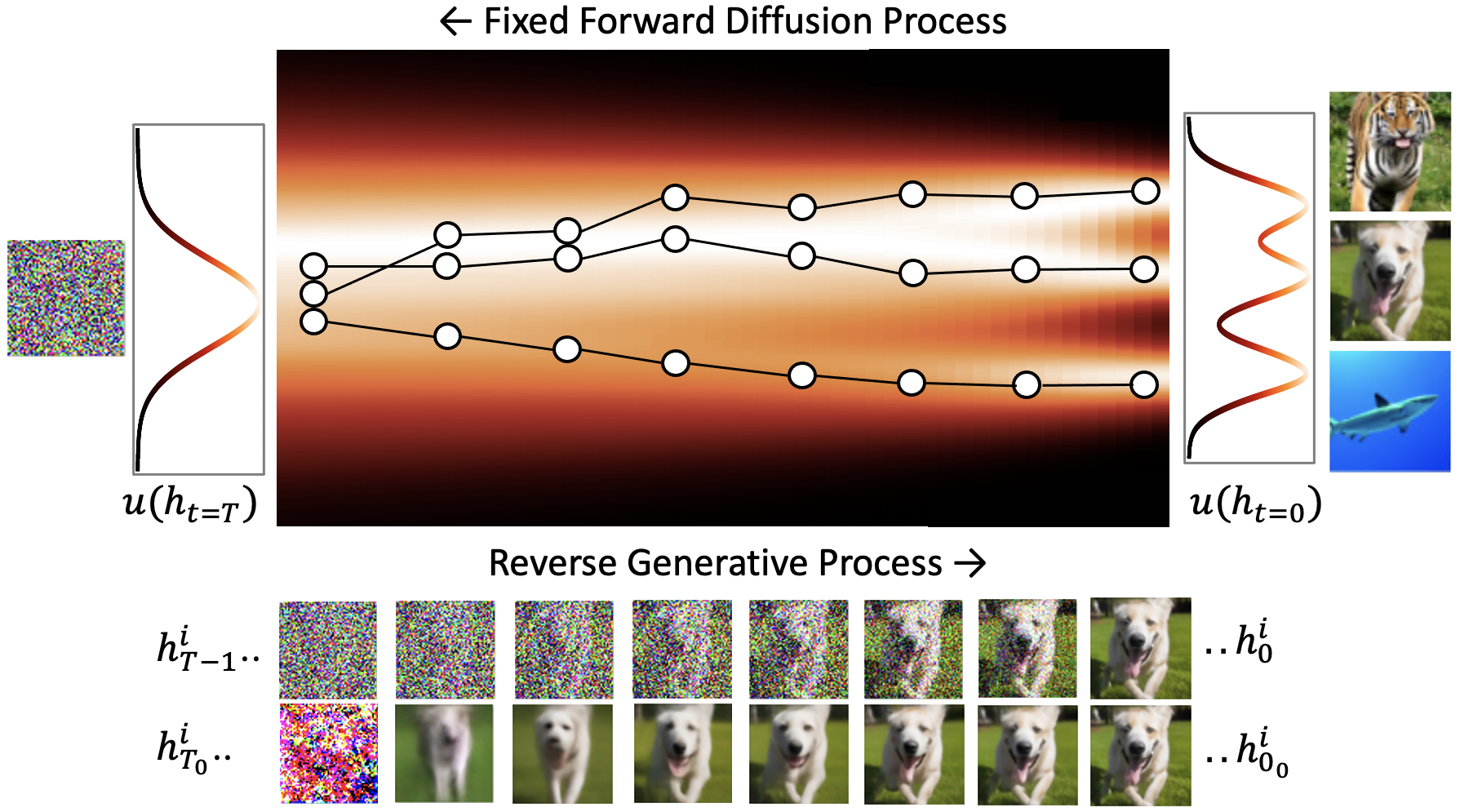}
    \caption{Theoretical representation of the diffusion process~\cite{song2021scorebased}, where the training data distribution $u(\textbf{h}^i_0)$ is represented as a 1D Gaussian. During generation (left to right), instances evolve from noise $\textbf{h}^{i}_T$ to a coherent image $\textbf{h}^{i}_0$\remove{, or a fuzzy denoised estimation $\textbf{h}^{i}_{t_0}$ of the final image at $t$ to a clear one.}\addition{. Intermediate steps also produce fuzzy, denoised estimates $\textbf{h}^{i}{t_0}$, which gradually refine into clearer images on convergence.}}
    \label{fig:diffusion_pdf}
\end{figure}
\textbf{Diffusion models}~\cite{sohl2015deep, ho2020denoising, nichol2021improved} aim to approximate the true distribution $u(\textbf{h}_0)$ of training data $\textbf{h}_0$. 
For this, training data is gradually corrupted with Gaussian noise, resulting in a sequence of data sets $\textbf{h}_1,\cdots,\textbf{h}_t,\cdots, \textbf{h}_T$ that define different spaces. 
The amount of noise added varies across iterations $t$, and is defined by a scheduler.
This corruption process of moving from the original data distribution $u(\textbf{h}_0)$ to noise $u(\textbf{h}_T)$, is called the forward process (see \cref{fig:diffusion_pdf}).

For generation, a single model $v_\theta$ (often a U-Net~\cite{unet}) is trained to estimate the noise needed to go from a noisy image instance $i$ at any iteration $t$, $\textbf{h}^i_t$, to the final denoised image $\textbf{h}^i_{0}$. 
We refer to this \remove{smooth estimate}\addition{\textit{step-wise denoised estimate}} of instance $i$ at iteration $t$ as $\textbf{h}^i_{t_0}$. 
A known noise level based on the scheduler is added back to $\textbf{h}^i_{t_0}$ to generate \addition{the \textit{step-wise noisy estimate}} $\textbf{h}^i_{t-1}$. 
The iterative process of estimating $\textbf{h}^i_{t-1}$ from $\textbf{h}^i_{t}$ continues until $t=0$ and is called the reverse generative process (see \cref{fig:diffusion_pdf}).  
Further details can be found in the original papers~\cite{ho2020denoising, nichol2021improved}. 
\remove{Although there have been several advancements, for example, skipping iteration steps to expedite sampling~\mbox{\cite{zhang2023gddim}}, the fundamental methodology of iterative reconstruction, where a less noisy image is predicted at each step, remains the same.}
For our analysis, we extract the noisy image $\textbf{h}^i_t$ and denoised estimate $\textbf{h}^i_{t_0}$ of the final denoised image at each iteration $t$ across instances $i$.
\newline
\newline
\noindent
\textbf{t-Distributed Stochastic Neighbourhood Embedding} or t-SNE~\cite{van2008tsne} is a non-linear DR method widely used for visualizing high-dimensional data. 
It aims to preserve the neighborhoods between the high-dimensional data points in a low-dimensional output space.
For this, t-SNE translates distances between the high-dimensional points as a symmetric joint probability distribution $P$. 
Similarly, a joint probability distribution $Q$ describes the low-dimensional similarity. 
The goal is to achieve a mapping or an embedding in the low-dimensional space $Q$ that faithfully represents $P$. 
This is achieved by optimizing the positions of the low-dimensional points via a cost $C$, which is the KL divergence between $P$ and $Q$,

\begin{equation}
\label{eq:tsnekl}
    C = KL(P||Q) = \sum^n_i \sum^n_{j, j \neq i} p^{ij} \ln\dfrac{p^{ij} }{q^{ij} } 
\end{equation}

$p^{ij}$ is the similarity between two high-dimensional points $\textbf{h}^i$ and $\textbf{h}^j$ of a data set $\textbf{h}$. It is based on their conditional probabilities $p^{i|j}$ and $p^{j|i}$, shown in \cref{eq:tsne_pij}. $p^{j|i}$ is the likelihood that $\textbf{h}^j$ is selected as the neighbor of $\textbf{h}^i$ based on a Gaussian probability density function centered at $\textbf{h}^i$ with variance $\sigma_i$. $\sigma_i$ is defined based on the local density of $\textbf{h}^i$ in the high-dimensional space and a defined perplexity parameter.

\begin{equation}
\label{eq:tsne_pij}
   p^{ij} = \dfrac{p^{i|j} + p^{j|i} }{2N}, \\
   where, \\
   p^{j|i} = \dfrac{\quad\quad exp(-||\textbf{h}^i - \textbf{h}^j||^2/(2\sigma^2_i))}{\sum_{k\neq i} exp(-||\textbf{h}^i - \textbf{h}^k||^2/(2\sigma^2_i))} 
\end{equation}

For the low-dimensional space $Q$, whose point positions need to be optimized, a student t-distribution with one degree of freedom is used to compute the joint probability distribution. The similarity $q^{ij}$ between two low-dimensional points $\textbf{l}^i$ and $\textbf{l}^j$ is then given by,

\begin{equation}
\label{eq:tsne_qij}
   q^{ij} = \dfrac{\quad\quad (1+||\textbf{l}^i-\textbf{l}^j||^2)^{-1}}{\sum_{k\neq m}(1+||\textbf{l}^k-\textbf{l}^m||^2)^{-1}} 
\end{equation}

Our paper employs the vanilla t-SNE~\cite{van2008tsne} to identify the similarity between high-dimensional images $\textbf{h}^i_{t}$ from the diffusion model. 
We augment t-SNEs' KL divergence cost function by introducing components to group images within an iteration $t$ and align images of an instance $i$ across iterations.\remove{While we use the basic t-SNE for simplicity of implementation, advanced versions~\mbox{\cite{pezzotti2019gpgpu, pezzotti2016approximated}} are compatible.} 

\section{Related Work}
\textbf{Interpreting Diffusion Models}: 
\remove{Previous works have extensively explored the capabilities of diffusion models~\mbox{\cite{yang2023diffusionsurvey}}. 
} While exploring and understanding the inner workings of diffusion models is important for leveraging their full potential and inspiring further research\cite{chang2023design}, their inherent complexity of high-dimensional evolutionary processes poses challenges~\cite{videau2023interactive, lee2023diffexplainer}, even for domain experts. 

Literature has shown the purpose of each iterative step, which correlates with the varying granularity of visual concepts~\cite{deja2022analyzing, si2023freeu, prasad2023unraveling, park2024timeconcepts}. 
Prior works have examined the local latent structure around an instance in the diffusion process~\cite{park2024riemannian, kwon2023diffusion} with methods like Riemannian geometry~\cite{park2024riemannian}.
While the goal was image editing, they also shed light on the granularity of features across steps~\cite{park2024riemannian}.
\remove{However, the evolution of the actual features and the decision-making dynamics of the model are missing. 
This distinction highlights the unique focus with $TDL$ on the holistic view of the evolutionary space and decision-making dynamics within diffusion models.}
However, it is crucial to understand the evolution of actual features to understand the root failure causes and enable downstream development~\cite{luccioni2023stablebias, park2024timeconcepts}.
Most of these works focus on applying insights to improved models rather than generic means of visualizing the evolutionary process to support understanding.

Pixel-level attribution maps explain relations between generated image areas and prompt keywords via cross-attention keyword–pixel scores in the denoising network of diffusion models~\cite{tang2023daam}. 
These maps have been used for controlling output images~\cite{hertz2022prompt, parmar2023zero, Tumanyan_2023_CVPR, chefer2023attend}.
Saliency map methods \remove{like GradCAM~\mbox{\cite{selvaraju2017grad} }}were extended to diffusion models to study the model's focus in each iteration and how it evolves~\mbox{\cite{park2024timeconcepts}}. 
However, these existing approaches primarily concentrate on individual samples, creating a gap in understanding the global evolutionary process, which is our focus. 

While visual analytics (VA) has provided a more generic means to analyze and understand complex DL models, most works focus, for example, on classification tasks~\cite{prasad2022transform, wang2020cnn, pezzotti2017deepeyes}, or generative adversarial networks~\cite{kahng2018gan}, which do not contain the iterative complexities of diffusion models. 
\addition{Methods to understand text-to-image data distributions have been proposed~\cite{zhao2024cupid}, but they primarily focus on analyzing final outputs rather than the full evolutionary process.} Lee et al.~\cite{lee2023diffexplainer} proposed DiffusionExplainer to provide insights into the full generative process of a diffusion model at an instance level. It projected intermediate noisy images via UMAP to a 2D space, offering insights into how individual instances evolve. However, our goal is to extend this to a holistic understanding of the entire evolutionary process. \addition{Visualizing the full evolutionary process offers deeper insights into how features emerge. This could help inform model design, for example, by revealing how schedulers impact generation informing model distillation~\cite{meng2023distillation} or identifying where feature entanglements occur, enabling more targeted adjustments~\cite{yang2023disdiff}.}
\newline
\newline
\noindent
\textbf{Visualizing high-dimensional evolutionary data}:
DR techniques, like t-SNE or UMAP, offer value in visualizing data evolution over a few instances~\cite{lee2023diffexplainer}.
Such methods have also been used in visualizing data evolution globally in strategic games~\cite{hinterreiter2021projection}. However, they do not preserve the iterative structure inherent in diffusion models due to their stochastic nature, which is amplified when visualizing large-scale samples. 
Neural network features across layers~\cite{rauber2016visualizing} or optimization iterations~\cite{li2020visualizing} have been visualized with one embedding per layer, making it challenging to study evolution.
While parallel coordinates are a popular means to visualize evolutionary data in other applications~\cite{fang2020survey, cantareira2021explainable}, projecting high-dimensional data onto a single dimension presents challenges. 
Luo et al.~\cite{luo2022dimenfix} incorporate soft Gaussian constraints to position points within an iteration close together. 
Methods using vertical distances between pairs of connected points can help with alignment of these points~\cite{pezzotti2018multiscale,cantareira2020generic}.
However, they overlook that early iterations in diffusion models contain less information than later ones, making equal visual space allocation inefficient for our case. 
\addition{Our evolutionary embedding addresses this by combining soft iteration displacement and instance alignment constrains and optimizes space usage across steps to better study diffusion model evolution.
}\remove{
In our evolutionary embedding, we aim to combine soft constraints for iteration displacement and instance alignment and extend it to the context of diffusion models. 
We further extend these concepts to a layout that optimizes space usage across generative iterations and facilitates the comprehensive study of diffusion model evolution.}

\section{\addition{EvolvED}}

\addition{The theoretical representation of data evolution in diffusion models~\cite{vahdat2021score, song2021scorebased} in \cref{fig:diffusion_pdf} shows how the data distribution $u(\textbf{h}_t)$ transitions from a 1D uni-modal Gaussian mode, representing pure noise, to a multi-modal distribution. 
Such a representation supports understanding the feature space evolution across iterations. 
However, translating this illustrative example to real high-dimensional, iterative generation spaces produced by real diffusion models is highly complex.
To address these challenges, we introduce $EvolvED$ (see \cref{fig:teaser}), a method that employs a structured pipeline to provide a holistic understanding of the iterative and high-dimensional generation process in diffusion models.
It begins by utilizing user-defined research questions to narrow the exploration of the vast generative space (see \cref{fig:teaser}a). 
$EvolvED$ allows for a targeted investigation into specific image attributes via tailored prompts aligned with the research question. 
Intermediate images are systematically extracted at multiple stages to capture the iterative progress and data evolution (see \cref{fig:teaser}b).
Given that diffusion models often lack a unified low-dimensional latent representation~\cite{prasad2022transform}, we propose employing feature extractors to isolate and track the attributes of interest throughout the generative process as shown in \cref{fig:teaser}c. 
Finally, to enhance the visualization of this iterative process, we introduce an evolutionary embedding algorithm in \cref{fig:teaser}d. 
This algorithm explicitly encodes iterations to enable the visualization of data evolution.
Each of these components of $EvolvED$ is described below.}

\subsection{User-Defined Focus \& Prompts} \addition{
    The $EvolvED$ pipeline starts by defining specific research questions to guide the exploration of the generative space (see~\cref{fig:teaser}a), such as, the $``$progression of natural objects$"$ to understand which attributes arise when. 
    With these research questions, $EvolvED$ aims to navigate the vast generative space. 
    Tailored prompts aligned with these questions must then be defined to ensure a focused analysis of relevant attributes. 
    For example, prompts like $``$a dog$"$, $``$a dog on grass$"$, or $``$a dog in water$"$ can help analyze the $``$progression of natural objects$"$.}

\subsection{Sampling Intermediate Data}
    \addition{To capture the model's iterative nature, we sample the generative space using tailored prompts, extracting intermediate outputs for an instance $i$ across multiple iterations $t$ (see~\cref{fig:teaser}b), rather than just the final output.
    This data includes the step-wise noisy $\textbf{h}^i_{t-1}$ and step-wise denoised estimates $\textbf{h}^i_{t_0}$.
    The denoised images represent the model’s prediction power of the final image at $t$, commonly transitioning from fuzzy object shadows to more detailed images (see~\cref{fig:diffusion_pdf}). $\textbf{h}^i_{t_0}$ is either explicitly predicted or can be implicitly extracted from diffusion models. 
    }

\subsection{Feature Extraction} 
\addition{
    Unlike classification models, which have well-defined latent layers for analysis, diffusion models use image-to-image architectures like U-Nets, which lack a unified low-dimensional latent space~\cite{prasad2022transform}. 
    The role of each layer shifts across iterations~\cite{prasad2023unraveling, prasad2022transform}, complicating the analysis of high-dimensional intermediate samples. 
    Latent space diffusion models~\cite{stablediff} use structured, high-dimensional spatial latent representations, with similar challenges to raw image analysis. 
    Hence, $EvolvED$ employs feature extractors $f$ specifically designed to isolate or encode features of interest from high-dimensional images to a lower-dimensional space. 
    These extractors process the intermediate outputs, where
    $\widehat{\textbf{h}}{}^i_{t-1}=f(\textbf{h}^i_{t-1})$ and $\widehat{\textbf{h}}{}^i_{t_0}=f(\textbf{h}^i_{t_0})$ represent the extracted features from the step-wise noisy and denoised estimates, respectively.
    For example, a robust ImageNet classifier $f$ encoding specific attributes like $grass$ and $dog$ would be effective in analyzing $``$progression of natural objects$"$. 
    The feature extractor chosen depends on the research question, as explored in our case studies. 
    While this selection is case-dependent and not the primary focus of our work, we provide guidelines for choosing effective extractors in different scenarios.}

    \begin{itemize}
        \item \addition{Correlation of Features and Insights: 
        Chosen feature extractors must either explicitly (classifiers~\cite{bai2021transformers}) or implicitly (encoders~\cite{clip}) extract features that align with the research question.}
        
        \item  \addition{Reliability on Corrupted Images: The feature extraction process must remain reliable even in the presence of corruption, such as fuzziness in intermediate images. 
        We evaluate different extractors on both noisy $\textbf{h}^i_{t}$ and denoised $\textbf{h}^i_{t_0}$ images to ensure that the features remain meaningful from the early stages of generation. }
    
        \item \addition{Smooth Transitions: Ideally, the extracted features should correspond to relevant changes in the images. If the images evolve gradually across iterations, the features should reflect this, and conversely, if the changes are abrupt, the features should capture that as well. This ensures a coherent representation of the instance's development throughout the diffusion process.}
        

    \end{itemize}
    \subsection{Visualization: Evolutionary Embedding}
   \addition{From the previous steps, we obtain one or multiple high-dimensional feature spaces per iteration.
   Existing DR methods to study high-dimensional data like t-SNE are not designed to preserve the evolutionary context while preserving semantic relations. Hence, we introduce a novel evolutionary embedding algorithm to address these challenges, which is the core $EvolvED$. 
   We consider common layout design choices, i.e., rectilinear and radial layouts~\cite{pandey2022genorec}, to explicitly represent iterations. 
   It allows the visualization of feature evolution while maintaining iterative context (see~\cref{fig:teaser}d).}
   Our evolutionary embedding method projects the  extracted features ($\widehat{\textbf{h}}{}^i_{t-1}$ or $\widehat{\textbf{h}}{}^i_{t_0}$) to a 2D space. \addition{The loss function is defined by three main components:}
   \remove{To achieve this, our method has the following goals:}

\begin{itemize}
    \item \textbf{Semantic loss (G1)}: Semantically similar elements in the high-dimensional feature space need to be placed together in the low-dimensional projection.
    \addition{This is achieved through the standard t-SNE KL divergence loss~\cite{van2008tsne} per iteration, which preserves the neighborhood structure of the high-dimensional data in the low-dimensional space. It is assumed that the feature space obtained by the previous modules is semantically meaningful. }
    \item \textbf{Displacement loss (G2)}: Elements belonging to the same iteration must be placed together to study and compare how different data modes evolve. 
    For example the topmost $shark$ mode in \cref{fig:diffusion_pdf} emerges sooner than the other objects.
    \addition{Hence, we impose a specific region per iteration of diffusion models using a displacement loss. 
    This loss uses a Gaussian centered around the corresponding iteration offset in the low-dimensional space (see \cref{fig:teaser}d).}

    \item \textbf{Alignment loss (G3)}: \remove{Pathways connecting elements of a specific instance across iterations should intersect minimally.
    This requirement ensures we only observe actual branching patterns and avoid misalignment across iterations due to the method's stochastic nature.}\addition{There should be alignment of components of the same sample $i$ across iterations $t$ to preserve coherence and minimize misalignments in the projections across $t$ due to the stochasticity of t-SNE.
    Hence, we use an alignment loss that measures the offset in vertical position~\cite{cantareira2020generic} or angular distance among consecutive pairs of elements of an instance across iterations.}
\end{itemize}

\begin{figure}
    \centering
    \includegraphics[width=0.99\linewidth]{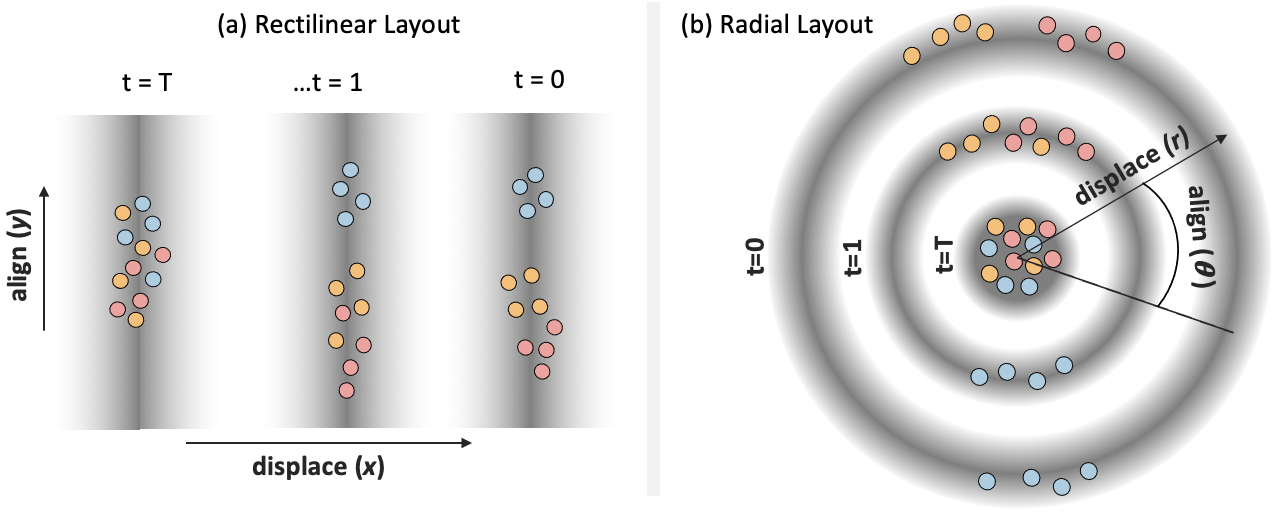}
    \caption{Evolutionary embedding process for the (a) rectilinear and (b) radial layout.
    Elements of iterations $t$ are attracted to dark, low-cost areas based on a Gaussian at a displacement value ($\bar{x}_t$ or $\bar{r}_t$) per $t$.
    Elements of an instance across iterations are aligned based on their $y$ or $\theta$.}
    \label{fig:gauss}
\end{figure}
To implement these goals, we define the total loss $C$ as a weighted combination of the three components, 
\begin{equation}
\label{eq:cost_all}
    C = \alpha*C_{s} + \beta*C_{d} + \gamma*C_{a}
\end{equation}
$C_s$, $C_d$, $C_a$ are the semantic (\textbf{G1}), displacement (\textbf{G2}), and alignment loss (\textbf{G3}) respectively.
The combined cost $C$ is minimized via gradient descent with a similar optimization method proposed by van der Maaten~\cite{van2008tsne}.
The displacement and alignment losses are additional constraints on each low-dimensional embedding.
\addition{The semantic t-SNE loss (\textbf{G1}) and displacement loss (\textbf{G2}) are local, i.e., they are applied within an iteration, while the alignment loss (\textbf{G3}) operates across steps by requirement. 
Since we want to preserve semantic clusters while maintaining the alignment and displacement conditions, the three losses must be modeled as a joint optimization problem with \cref{eq:cost_all}.}
\newline
\newline
\noindent
\textbf{Common t-SNE loss (G1)}: For both layouts, the standard t-SNE loss (\textbf{G1}) is applied separately for each iteration $t$ to preserve semantic neighborhoods from the high-dimensional to the low-dimensional space per $t$. Specifically, \cref{eq:tsnekl} is used to compute the \textbf{G1} loss, denoted as $C_s$, for each iteration independently as follows. 
\begin{equation}
\label{eq:tsne_t}
\begin{aligned}
    C_{s} &= \dfrac{1}{T*N} \sum_t \text{KL}(P_t||Q_t) = \sum_t \sum_i \sum_{j, j \neq i} p^{ij}_t \ln\frac{p^{ij}_t }{q^{ij}_t } \\
\end{aligned}
\end{equation}
$C_s$ is normalized by the data size ($N$ instances and $T$ samples).
Gradients $\partial C_s/\partial x^i_t$ and $\partial C_s/\partial y^i_t$ are computed as in the original paper~\cite{van2008tsne}.
\newline
\newline
\noindent
\textbf{Rectilinear Layout:} models each iteration as a vertical line (see \cref{fig:teaser}d). 
The low-dimensional elements \( \mathbf{l}^i_t \), with coordinates \( (x^i_t, y^i_t) \), are arranged along the x-axis according to their iteration $t$ with the Cartesian displacement loss $C{}^c_d$ (\textbf{G2}). 
$C{}^c_d$ is designed as a Gaussian centered around an x-coordinate offset \( \overline{x}_t \) with a thickness determined by the Gaussians' standard deviation \( \sigma \) for each $t$ in the low-dimensional space. 
\begin{equation}
\label{eq:rect_gauss}
    C{}^c_d =  \dfrac{1}{T*N}\sum_t\sum_i C^c_d(i, t) \text{, where, } C^c_d(i, t) - \dfrac{1}{\sigma\sqrt{2\pi}} e^{-\dfrac{(x^i_t-\overline{x}_t)^2}{2\sigma^2}}
\end{equation}

The Gaussian attracts elements at iteration $t$ to \( \overline{x}_t \), hence explicitly encoding iterations (see \cref{fig:gauss}a).
Initially, \( \sigma \) is set to a larger value \( \sigma_{start} \) to allow for more freedom in point movement, then reduced gradually to \( \sigma_{end} \) across optimization iterations to bring points closer to their corresponding \( \overline{x}_t \).
Vertical lines representing different iterations are positioned with a buffer spacing $s$ between them. 
Noisy iterations are positioned on the left side.

The Cartesian alignment loss $C{}^c_{a}$ (\textbf{G3}), aligns elements of each instance $i$ across iterations.
It is defined as the vertical distance between pairs of the same instance $i$ across iterations consecutive $t$ pair (see \cref{eq:rect_align}) as in literature~\cite{pezzotti2018multiscale, cantareira2020generic}.
Larger $t$ values correspond to early generation steps.
\remove{For the rectilinear layout, empirical findings suggest setting 
$\alpha=1, \beta=1, \gamma=0.2$, since these values consistently yield desired results.
}The cost and gradients are computed as follows,

\begin{equation}
\label{eq:rect_align}
    C{}^c_{a} = \dfrac{1}{N*T}\sum_i\sum^{T}_{t=1} C^c_a(i, t)\text{, where, } C^c_a(i, t) = ||y^i_t-y^i_{t-1}||^2
\end{equation}

\begin{align}
\label{eq:rect_grad}
\begin{aligned}[t]
    \frac{\partial C}{\partial x^i_t} &= \alpha*\frac{\partial C_{s}}{\partial x^i_t} + \beta*\frac{\partial C{}^c_{d}}{\partial x^i_t}\\
    \frac{\partial C{}^c_{d}}{\partial x^i_t} &= \frac{(x^i_t-\overline{x}_t)}{2\sigma^2}*C{}^c_{d} (i, t)\\
\end{aligned}
\quad
\begin{aligned}[t]
    \frac{\partial C}{\partial y^i_t} &= \alpha*\frac{\partial C_{s}}{\partial y^i_t} + \gamma*\frac{\partial C{}^c_{a}}{\partial y^i_t}\\
    \frac{\partial C{}^c_{a}}{\partial y^i_t} &= \frac{(y^i_t-y^i_{t+1})}{C{}^c_{a} (i, t)}\\
\end{aligned}
\end{align}

\remove{Where, $X^{ij}_t = (1+(x^i_t-x^j_t)^2)^{-1}$, and, $Y^{ij}_t = (1+(y^i_t-y^j_t)^2)^{-1}$.
While $C_{\text{s}}$ in \mbox{\ref{eq:tsne_t}} illustrates the high-dimensional elements $\textbf{h}^i_t$, these elements can also be other high-dimensional elements such as $\widehat{\mathbf{h}}{}^i_t$.}
\addition{The resultant force exerted on the data point $\mathbf{l}^i_t$ by all three costs ($\partial C/\partial x^i_t$ and $\partial C/\partial y^i_t$) causes the points to move within the vertical band corresponding to iteration $t$.
}\remove{
$\partial C/\partial x^i_t$ and $\partial C/\partial y^i_t$ are the resultant force exerted on
the data point $\mathbf{l}^i_t$ by all three costs. 
We simultaneously optimize the three loss functions.
The resultant force of the three costs leads to points moving within the vertical band corresponding to iteration $t$. }
We initialize $\mathbf{l}^i_t$ in a buffer region around the corresponding $\overline{x}_t$ for faster optimization results.
The expected result is that points are clustered based on semantic feature similarity around their corresponding x-offsets $\overline{x}_t$ per iteration while maintaining the instance alignment across iterations (see \cref{fig:teaser}).
\newline
\newline
\noindent
\textbf{Radial layout:} The rectilinear layout fulfills our goals by enhancing the visualization of branching patterns and resembling the theoretical representation of expected behavior in \cref{fig:diffusion_pdf}. 
However, it uses space inefficiently, as all iterations are allocated equal space, even though early-generation steps, which are mostly noise, contain less relevant patterns.
To address these gaps, we propose a radial layout with concentric rings (see \cref{fig:teaser}d and \cref{fig:gauss}b). 
Each iteration is represented as a ring; innermost rings show early noisy iterations and the outer rings show refined steps. 
\remove{A key difference between the rectilinear and radial layout is the transformation of the cost function to use the perspective of polar coordinates.
In a radial layout, early noisy iterations occupy less space at the center compared to later iterations in the outer rings, addressing sub-optimal space utilization.}
\addition{Unlike the rectilinear layout, the radial layout optimizes space by assigning less area to early iterations and more to later ones. It also allows points to move flexibly around the rings, which may result in more effective embeddings.}

The low-dimensional elements $\textbf{l}^i_t$ are represented in polar coordinates ($r^i_t$, $\theta^i_t$) to facilitate the definition of forces in the radial layout. 
Conceptually, the same losses $C_{\text{s}}$, $C^c_{\text{d}}$, and $C^c_{\text{a}}$ are utilized, but the form is adapted to the radial layout. 
To preserve correctness of the Euclidean distance gradients in $C_{\text{s}}$, they are computed in Cartesian space and then converted to polar coordinates.
The displacement loss ($\textbf{G2}$) was adapted for the radial layout $C{}^p_{d}$ to organize elements of an iteration along the rings.   
\remove{This is modeled as a Gaussian centered around a radius-offset $\overline{r}_t$ per iteration in the embedding as defined in \mbox{\ref{eq:rad_gauss}}. 
The Gaussian encodes iterations by attracting elements at iteration $t$ toward the ring \mbox{ $\overline{r}_t$} (see \mbox{\ref{fig:gauss}}).
}\addition{It is modeled as a Gaussian centered around a radius-offset $\overline{r}_t$ per iteration in the low-dimensional space in \cref{eq:rad_gauss}. }
\begin{equation}
\label{eq:rad_gauss}
    C{}^p_{d} = \dfrac{1}{T*N} \sum_t\sum_i C^p_d(i, t) \text{, where, } C^p_d(i, t) = - \dfrac{1} {\sigma\sqrt{2\pi}} e^{-\dfrac{(r^i_t-\overline{r}_t)^2}{2\sigma^2}}.
\end{equation}
The Gaussian attracts elements at iteration $t$ toward the corresponding ring center, $\overline{r}_t$ (see \cref{fig:gauss}b).
The innermost ring corresponds to $\overline{r}_{T}=0$, forming a circle.
Subsequent rings have user-defined offsets $\overline{r}_{t!=T}>0$, which determine the spacing $s$ between iterations as before.
\remove{This spacing was set to $20$, i.e., $\overline{r}_{t}=20*(T-t)$. }

The radial alignment loss (\textbf{G3}), $C{}^p_{a}$, ensures that elements of the same instance $i$ across iterations are aligned, and it is defined as:
 \begin{equation}
\label{eq:rad_align}
    C{}^p_{a} = \dfrac{1}{N*T} \sum_i\sum^{T}_{t=1} 1 - \left| \text{sim}^i_t \right|, where, \text{sim}^i_t = \cos\left(\frac{{\theta^i_t - \theta^i_{t-1}}}{2}\right).
\end{equation}
We minimize the cosine distance between the angles $\theta^i_t$ of sequential pairs of corresponding elements. 
Halving the angular difference ensures the maximum cost is at $\pi$.
We use the absolute value of cosine similarity to prevent poor local minima, particularly at 
$\pi$, where the angular difference is largest but the gradient of a cosine function is nearly flat. This could slow convergence when similar points are stuck on opposite sides. 
\remove{The final loss $C{}^p$ is the weighted sum of $C_s$, $C{}^p_d$, and $C{}^p_a$ with respective weights $\alpha$, $\beta$, and $\gamma$.}The optimization process follows a similar approach as the rectilinear one. 
However, in this case, the resultant force of the three cost elements leads to points moving along the respective iteration rings.
The gradient updates in polar form are defined as follows:
\begin{align}
\label{eq:rad_grad}
\begin{split}
    \frac{\partial C{}^p_d}{\partial r^i_t} &= \frac{(r^i_t-\overline{r}_t)}{\sigma^2}*C{}^p_d(i, t)\\
    \frac{\partial C{}^p_a}{\partial \theta^i_t} &= 
    \begin{cases}
    -\frac{1}{2}\sin\left(\frac{\theta^i_t - \theta^i_{t-1}}{2}\right) & \text{if } \text{sim}^i_t < 0 \\
    \frac{1}{2}\sin\left(\frac{\theta^i_t - \theta^i_{t-1}}{2}\right) & \text{if } \text{sim}^i_t > 0
    \end{cases}
\end{split}
\end{align}

\addition{The gradient of 
$\partial C{}^p_a/\partial \theta^i_t$ is undefined when $\text{sim}^i_t = 0$, where the point can move in either direction along the ring. While very unlikely to occur, we choose one of the two possible gradient directions randomly.}
\remove{The gradient of $C{}^p_{a}$ with respect to $\theta^i_t$ when $\text{sim}^i_t = 0$ is undefined. This is the unstable equilibrium point, where the point can move in either direction along the ring to reach a similar point that is diagonally opposite. While it is unlikely for $\text{sim}^i_t = 0$ in practice, one of the other two gradient options is randomly chosen during optimization.}

\addition{Note that the semantic $C_s$ and displacement $C_d$ in both layouts are inherently normalized as they are probability density functions (PDF). In rectilinear layouts, the vertical alignment loss $C^c_a$ is min-max normalized per iteration. The angular alignment loss $C^p_a$ in radial layouts is a scaled cosine loss $\in [0, 1]$, but becomes unbounded in Cartesian space. Since our optimization uses a fixed $\overline{r}_{t}$ across $t$ (later stretched for visualization), the costs remain consistent across datasets\footref{supp}. While $C^c_a$ and $C^p_a$ are not PDFs, such joint optimizations are commonly performed in literature~\cite{cantareira2020generic, pezzotti2018multiscale}, where variations in scales are managed through the weighting term; we follow this approach.}

\addition{ 
The choice of layout depends on the analytical objectives~\cite{pandey2022genorec}, with both offering distinct advantages and drawbacks. 
The rectilinear layout allocates equal space to all iterations, but not all layers have the same content in information. Early iterations often contain noisy, unstructured data, making it an inefficient use of space. 
In contrast, the radial layout optimizes space by allocating more room to later iterations, where more complex patterns emerge. 
While rectilinear layouts are useful for representing iterations, the main limitation, especially for representing semantic relations, is that elements at the extremes of the 1D-like embeddings can only show similarity in one direction (inward).
The radial layout allows for movement along opposite directions of the cycle, so similar points can spread more naturally and easily across the ring. 
Further, the continuous, cyclic nature of this radial layout avoids extremes, offering a uniform interpretation and gradual evolution of all elements. Both layouts can provide complementary insights and may be analyzed side by side, depending on the application.}

\subsection{Interactive Front-End}
As a proof-of-concept, we developed an interactive front-end\footnote{See supplementary materials for more details.\label{supp}} to explore data evolution within $EvolvED$.
The evolutionary embedding is visualized in either radial or rectilinear layouts. 
\remove{Iterations are represented as either concentric rings or vertical lines, highlighting progression from noisy (innermost ring or left-most vertical line) to clearer representations (outermost ring or right-most vertical line).
}When visualizing these embeddings to enable analysis, several considerations arise: 1) linking the embedding and the high-dimensional image space; 2) supporting analysis of the evolution of instances; and 3) characterizing images through prompt keyword-image links to improve comprehension of high-dimensional evolutionary data. 

To comprehend the high-dimensional image space via the embedding, we display images of specific iterations ($\textbf{h}^i_{t-1}$ or $\textbf{h}^i_{t_0}$) at regular intervals along the largest ring\remove{ or along the length of the right-most vertical line}, spaced at intervals of $\theta$ degrees\remove{ or $y$ units, respectively (see \mbox{\ref{fig:teaser}}d)}. 
The high-dimensional images $\textbf{h}^i_{t-1}$ or $\textbf{h}^i_{t_0}$ of selected embedding elements are dynamically displayed for a linked analysis. 
To further explore the evolution and hierarchy of data, keywords or attributes within a prompt are utilized to color the low-dimensional embedding points. 
For example, a prompt $``$a dog on grass$"$ can be colored by the keyword $dog$ or $grass$, analyzing the effect of each keyword on the evolution process. 
To enhance the visualization of instance evolution, pathways connecting images of the same instance across iterations are traced with smooth curves. 
Given the potential for visual clutter arising from showing all pathways, we visualize only the evolution of selected elements $\textbf{l}^i_{t-1}$ of a specific instance $i$ across iterations $t$ (see \cref{fig:imgnetpathways}). 
\remove{These pathways are drawn as smooth cardinal splines using respective elements $\textbf{l}^i_{t}$ of an instance $i$ as control points. 
The embedding points are clustered per iteration and keyword using DBSCAN~\mbox{\cite{dbscan}} to achieve a cleaner visualization of pathways.  
Pathways and points within each cluster are interpolated towards the cluster centroid based on a controllable interpolation factor.
While there could be better approaches to achieve this, we leave it to future work.}
Simple data filters on the iteration step $t$, path lengths, and prompt keywords are provided to facilitate analysis. 
Users can interactively control keyword-color associations and pathway interpolation factors.
It is important to note that our contribution primarily focuses \remove{on dimensionality reduction}\addition{on the $EvolvED$ pipeline} and evolutionary embedding rather than the interactive front-end aspect.
We leave it to future work to achieve a complete visualization design.

\section{Experiments with EvolvED components} 
Within \remove{$TDL$}\addition{$EvolvED$}, several components interact, including \addition{the raw (intermediate)} images\remove{ to be encoded}, feature extractors, \remove{the two layouts, and their loss metrics}\addition{the evolutionary embedding and its loss functions.
Understanding these interactions is crucial to elucidate how each component affects the outcomes of \textit{EvolvED} and how it can be optimized for various cases. 
The below sections detail a simple example, the impact of the feature encoders, and adding constraints to t-SNE with our evolutionary embedding method.}
\remove{Understanding how these components interact and their effects on evolutionary embedding within $TDL$ is crucial.
In the following sections, we present our setup utilized for these experiments. 
We then evaluate and compare the two proposed layouts with the baseline vanilla t-SNE. 
Finally, we explore various image encoders and analyze the two types of diffusion images we extract: smooth and noisy representations.
By doing so, we aim to gain insights into how each component influences the evolutionary embedding process within $TDL$ and aids in tuning the embedding quality for various datasets and applications.}

\subsection{Setup}
\label{sec:imagenetdata}
We use GLIDE~\cite{glide}, a popular text-to-image diffusion model, for our experiments.  
We generate a set of $1000$ samples from $10$ randomly selected ImageNet classes as prompts $\in [$tiger, bird, car, fruit, building, dog, shark, ball, plane, table$]$.
GLIDE's base model was used to generate samples in $100$ iterations. 
For the evolutionary embedding, $11$ equally spaced iterations were used. 
Empirical findings suggested setting 
$\alpha=1, \beta=1$ and $\gamma=0.05$, and $\gamma=0.2$ for the rectilinear and radial layouts consistently yielded desired results. 
We use \( \sigma_{start} \)$=20$, \( \sigma_{end} \)$=10$ out $\overline{x}_t=\overline{r}_t=20$, for optimization, stretched out with $s=20$ for visualization across all our cases\footref{supp}.
We examine the impact of the interim images ($\textbf{h}_{t-1}$, $\textbf{h}_{t_0}$) encoded with different image encoders $f$ to generate \addition{corresponding} intermediate spaces\remove{ of $\widehat{\textbf{h}}{}_{t-1}$ and $\widehat{\textbf{h}}{}_{t_0}$ respectively}.
The evolutionary embedding is applied to this intermediate space and the raw images for comparison.
The encoders explored include a robust ImageNet classifier~\cite{bai2021transformers} and the image encoder of CLIP~\cite{clip}.
CLIP is a foundational model that learns visual concepts from natural language supervision.

\subsection{Impact of Image Encoders \& Images Encoded}
\begin{figure}[!b]
    \centering
    \includegraphics[width=0.99\linewidth]{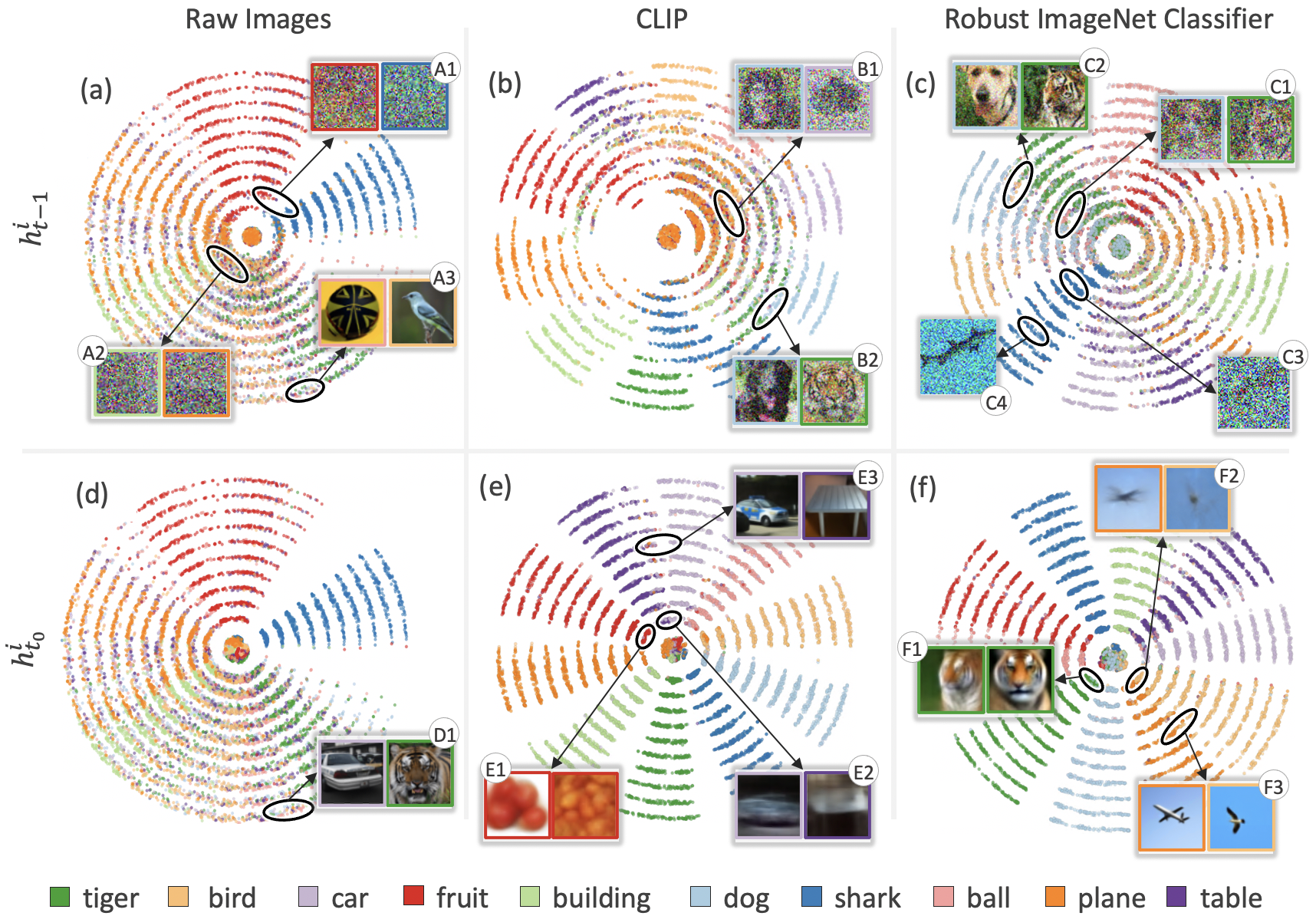}
    \caption{Applying evolutionary embedding on GLIDE images with various feature extractors. (a) Noisy $\textbf{h}_{t-1}$ and (d) step-wise denoised images $\textbf{h}_{t_0}$ undergo evolutionary embedding. The noisy $\textbf{h}_{t-1}$ are encoded with (b) CLIP image encoder, and (c) a robust ImageNet classifier~\cite{bai2021transformers} before the embedding creation. Similarly, denoised $\textbf{h}_{t_0}$ are encoded with (e) a CLIP and (f) a robust ImageNet classifier.
}
    \label{fig:encoders}
\end{figure}
\addition{

We evaluate the feature extractors affect $EvolvED$ in generating embeddings from intermediate images of the diffusion process. Our goal is to assess the encoders' effectiveness in extracting relevant semantic attributes while reflecting the inherent fuzziness of the images. We expect high levels of fuzziness in the high-dimensional images to result in mixed clusters across prompts in the early steps of the embedding, while later steps should produce distinct clusters as images improve.

Applying t-SNE to raw step-wise noisy (see \cref{fig:encoders}a) and denoised images (see \cref{fig:encoders}d) is a baseline for evaluating feature extractors. 
\Cref{fig:encoders} A1 shows that the noisy image colors can differ significantly at a pixel level, allowing their distinction in the embedding. However, with more nuanced semantic differences, raw image embeddings struggle to capture meaningful separation. When there are clearly semantic differences in images of birds, balls, or tigers in A3 and D1 (see \cref{fig:encoders}), the raw image embeddings fail to differentiate these due to their reliance on pixel-level information. This shortcoming highlights the necessity of a feature extractor that go beyond raw pixel data to extract high-level semantic features for meaningful analysis.

We evaluate two feature extractors: a robust task-specific classifier trained on ImageNet~\cite{bai2021transformers} and the foundational, generic CLIP model~\cite{clip}, which is not explicitly trained to be robust.
Both extractors provide more useful groupings in the embeddings compared to the raw image baseline in \cref{fig:encoders}. 
Comparing the two, we observe that CLIP~\cite{clip} struggles with noisy images in the early stages, as seen in \cref{fig:encoders} B1, where random images of dogs and cars are grouped together. 
However, by the final 30-40\% of iterations, CLIP shows more useful groupings in B2, such as between tigers and dogs with a green background.
In contrast, the robust classifier~\cite{bai2021transformers} excels even in the early-generation steps by reflecting image-level changes. 
For instance, \cref{fig:encoders} C1 shows that the classifier correctly groups dogs and tigers in a green background together in the early stages, reflecting their shared contextual elements, whereas CLIP struggles to do the same.
As the diffusion process progresses, the robust classifier maintains this contextual separation, resulting in well-defined clusters in \cref{fig:encoders} C2. Furthermore, in the very early stages (within the first 10\% of iterations), the robust classifier differentiates images based on broad color differences, such as between the blue shark and red fruit classes. 
This confirms that a robust feature extractor works effectively even in the initial iterations of the diffusion process.
While CLIP is not as effective as a robust classifier in the early stages, it still performs far better than the raw image baseline. This demonstrates the value of even a general-purpose extractor for noisy images, though robust models clearly offer an advantage.

When it comes to the step-wise denoised predictions at each iteration, the embeddings become clearer and more interpretable. 
Both the robust classifier (\cref{fig:encoders}f) and CLIP (\cref{fig:encoders}e) reflect the similarity between images in their embeddings. 
For example, when images are fuzzy in \cref{fig:encoders} E2 and F2, the embeddings reflect this with overlapping clusters, and they eventually separate these images as they become more distinct (E3 and F3). 
Images with distinct properties from the start, such as those in \cref{fig:encoders} E1 and F1, are separated early on. 
This demonstrates that both robust and general extractors work effectively on the step-wise denoised images, capturing the underlying semantic meaning.

In summary, robust feature extractors~\cite{bai2021transformers} work better for analyzing noisy images, while non-robust extractors~\cite{clip} perform equally well with intermediate, denoised images. The choice of encoder is crucial for embedding quality; mismatches between the feature extractor and prompt focus can lead to ineffective embeddings. For example, an ImageNet classifier would not be ideal for analyzing medical images. While general-purpose models~\cite{clip} outperform raw images, case-specific feature extractors, if available, are always preferable. 
}

\subsection{Evaluating Evolutionary Embedding}
\begin{figure}
    \centering
    \includegraphics[width=0.99\linewidth]{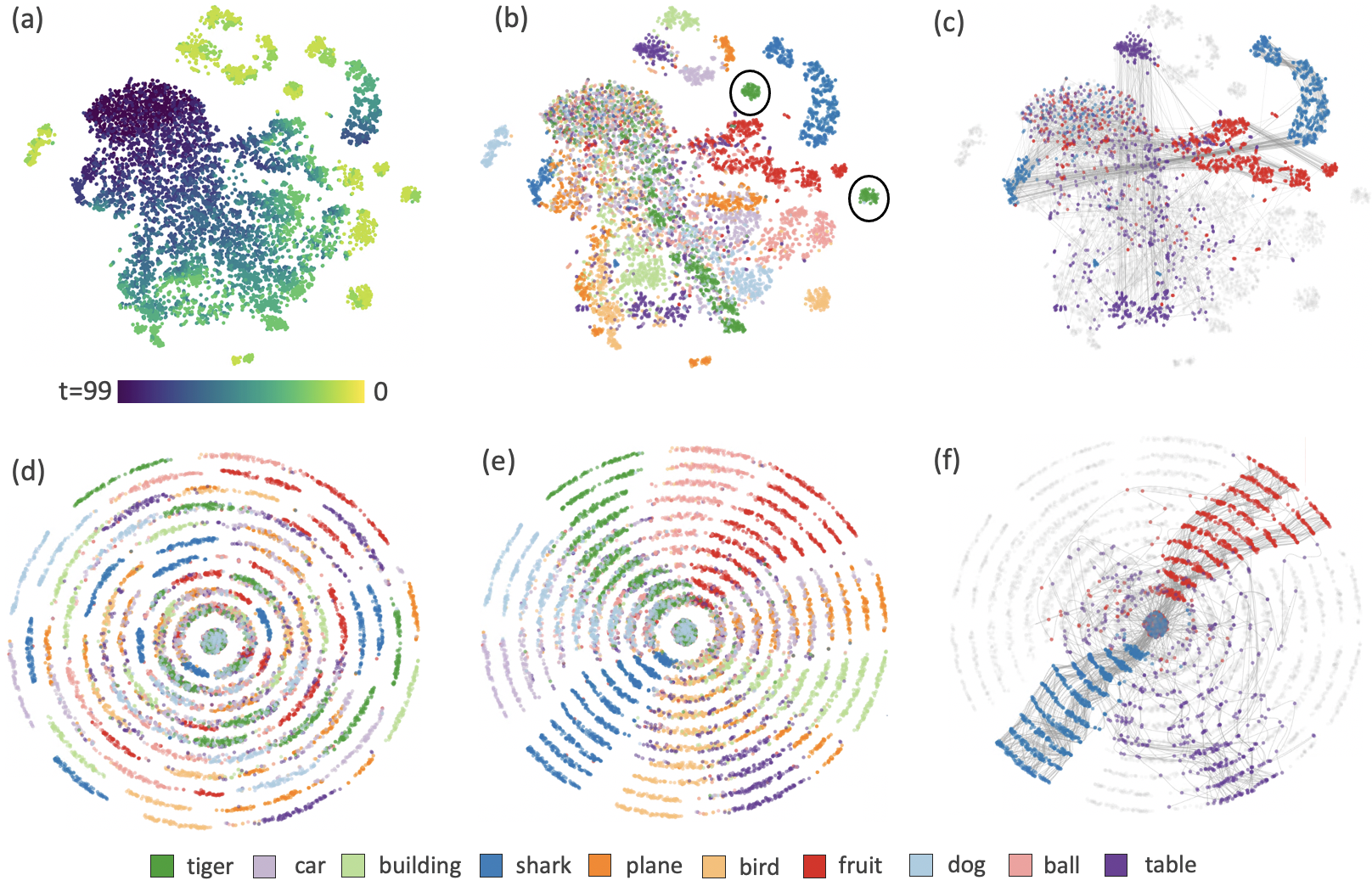}
    \caption{Single vanilla t-SNE~\cite{van2008tsne} embedding on all $\widehat{\textbf{h}}{}_{t-1}$ extracted from the GLIDE ImageNet use case with a classifier~\cite{bai2021transformers}. Points are colored by (a) iteration $t$ and (b) prompt. 
    Radial layout without (d) and with (e) the instance alignment loss. 
    Pathways of $``$tables$"$, $``$sharks$"$, $``$fruits$"$ are shown on the vanilla t-SNE (c) and radial layout (f).
}
    \label{fig:metric_utility}
\end{figure}
We explore each loss function \cref{eq:cost_all}, of our evolutionary embedding in \cref{fig:metric_utility} and compare the proposed layouts with \remove{vanilla t-SNE~\mbox{\cite{van2008tsne}}.
}\addition{with two baselines: 1) global vanilla t-SNE, embedding all steps $t$ in a single space, and 2) step-wise vanilla t-SNE, creating separate embeddings for each $t$, i.e, $C_s$ alone.} We create embeddings of the step-wise noisy estimates $\textbf{h}^i_{t-1}$ encoded with an ImageNet classifier~\cite{bai2021transformers} for evalutation.

\remove{This intermediate space is analyzed with our evolutionary embedding method and vanilla t-SNE, each optimized for $2000$ iterations.
The vanilla t-SNE, obtained by projecting all images on a single embedding, shows some form of evolution seen in the continuity of points in \mbox{\ref{fig:metric_utility}a}. 
However, as clusters become more distinct, e.g., the yellow points of \mbox{\ref{fig:metric_utility}}a, we lose the information about the progress through the iterations $t$, and about relations within iterations themselves (see blue points in \mbox{\ref{fig:metric_utility}}c). }
\addition{Global vanilla t-SNE enables analysis across steps, but as clusters become distinct, i.e., the yellow points of \cref{fig:metric_utility}a, we lose the information about the progress through the iterations $t$, and about relations within iterations themselves.
This makes it difficult to trace data evolution. Step-wise vanilla t-SNE offers better information within each iteration, but cluster evolution over time is difficult to trace, since each $t$ is independent.}
The non-aligned iterations and the stochastic cluster placement complicate the tracing of data evolution.
Our proposed radial layout in \cref{fig:metric_utility}e explicitly encodes iterations and uses spatial coherence for analysis of cluster evolution patterns\addition{, something neither baseline achieves}.
For example, the early evolution of $``$shark$"$ and $``$fruit$"$ in \cref{fig:metric_utility}f is not obvious in the global vanilla t-SNE in \cref{fig:metric_utility}a and b.
The radial layout (see \cref{fig:metric_utility}d) reduces stochastic cluster placement \remove{with the instance alignment loss }(see \cref{fig:metric_utility}e), while preserving cluster relations.

\remove{The evolutionary embedding aims to introduce aligning constraints for visual clarity of the diffusion process while preserving the high-dimensional relations within an iteration.
Hence, the performance of our method against the vanilla t-SNE~\mbox{\cite{van2008tsne}} in \mbox{\ref{fig:metric_utility}} to ensure preservation of the high-dimensional neighborhoods. 
We test this with projection quality metrics,}
\addition{To evaluate the extent to which our added displacement and alignment constraints preserve high-dimensional neighborhoods, we compare our evolutionary embedding to the baselines using projection quality metrics,} trustworthiness ($Q_\text{trust}$) and continuity ($Q_\text{cont}$)~\cite{kaski2003trustworthiness} per iteration $t$.
These metrics take values in [0,1]; one being the best performance.
$Q^t_\text{trust}$ checks whether the $k$ nearest neighbors of an element remain neighbours in the high-dimensional space within an iteration\addition{. }\remove{, where $Q{}^t_\text{trust}=1-A_k \sum_{i} \sum_{x^j_t \in U_k (x^i_t)} r(x^i_t, x^j_t) - k $.
Here, $r(x^i_t, x^j_t)$ signifies the rank of a sample $x^t_j$ concerning its distance to $x^i_t$ in the high-dimensional space. 
$U_k(x^i_t)$ is the set of data elements that are neighbors of $x^i_t$ in the embedding but not in the high-dimensional space.
$A(k)$ is a scaling factor.
Similarly, }$Q{}^t_\text{cont}$ checks whether the $k$ nearest neighbors of an element in the high-dimensional space is preserved in the embedding\addition{.}\remove{, where $Q{}^t_\text{cont}=1-A_k \sum_{i} \sum_{x^j_t \in V_k (x^i_t)} \hat{r}(x^i_t, x^j_t) - k $.
Here,  $V_k(x^i_t)$ are those elements that are neighbors of $x^i_t$ in the high-dimensional but not the embedding space.
} We utilize $k=7$ neighbors inline with previous works~\cite{meng2023class, kaski2003trustworthiness}.

\begin{figure}[!t]
    \centering
    \includegraphics[width=0.95\linewidth]{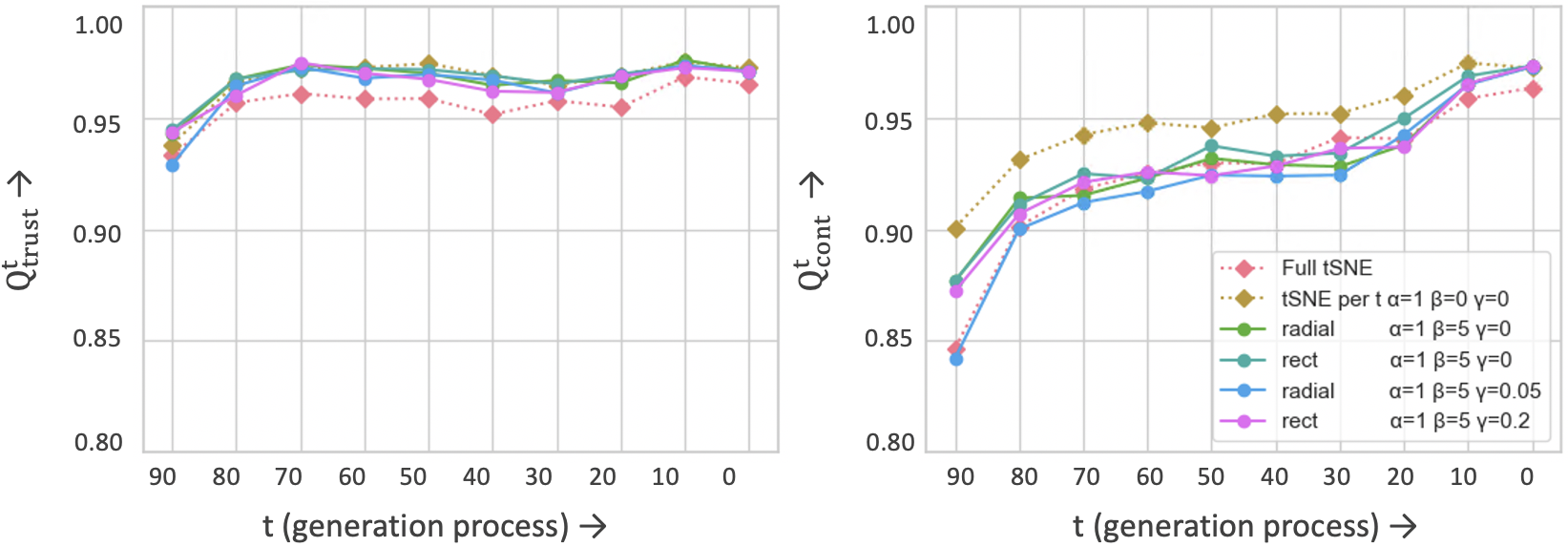}
    \caption{$Q^t_\text{trust}$ (left) and $Q^t_\text{cont}$ (right) of \addition{full} vanilla t-SNE \addition{across $t$, step-wise vanilla t-SNE per $t$ ($C_s$)}, and the proposed radial ($radial$) and rectilinear ($rect$) layouts without (\addition{$C_s$ with $C_d$}) and with the \remove{instance} alignment metric (\addition{$C_s$, $C_d$, and $C_a$}) across iterations $t$.
    Embeddings are obtained on $\widehat{\textbf{h}}{}^i_{t-1}$.
    \remove{$t=100$ is noise.
    $\alpha=1$.}}
    \label{fig:metric_acrosst}
\end{figure}
\remove{\mbox{\ref{fig:metric_acrosst}} illustrates $Q{}^t_\text{trust}$ and $Q{}^t_\text{cont}$ across various layouts. 
}$Q{}^t_\text{trust}$ and $Q{}^t_\text{cont}$ for the proposed evolutionary embeddings remain comparable to \addition{full} vanilla t-SNE in \mbox{\ref{fig:metric_acrosst}}.
\addition{While $Q{}^t_\text{trust}$ of per step vanilla t-SNE is similar to our layouts, there is a minor deterioration in $Q{}^t_\text{cont}$ in our layouts, primarily in early noisy steps, though they align better in later steps.
This means that while the neighbors our layout shows are true, all neighbors cannot be preserved.
This is expected, as step-wise vanilla t-SNE offers full 2D flexibility, while our method is constrained by a (narrow) Gaussian width per iteration, limiting how neighbors can be preserved.}
Further, there is minimal deterioration in both metrics when the alignment loss ($C_{a}$) is introduced \cref{fig:metric_acrosst}\remove{ compared to the metric without the alignment.
This }, indicating that our method preserves high-dimensional structure while offering clearer insights into data evolution \addition{when compared to both baselines.}
\remove{Additionally, the evolutionary embedding method significantly outperforms vanilla t-SNE in terms of speed for this experiment.
This is primarily because we divide t-SNE into smaller segments per iteration and combine them with simple alignment losses. 
The preprocessing step with PCA takes $\approx1$ minutes for our method in comparison to $\approx6$ minutes for the vanilla t-SNE.
Further, our method takes $<1$ second, while vanilla t-SNE takes $\approx 9$ seconds per iteration. 
This substantial reduction in computational time combined with the visual clarity of data evolution highlights the efficiency and effectiveness of our approach.
}Further, cluster quality metrics are comparable between the proposed radial and rectilinear layouts in \cref{fig:metric_acrosst}. 
\addition{The main decline in quality metrics of the proposed radial method comes in the largest $t$ value early in the generation process. 
Since these represent almost pure noise, so there is a lack of structure, the measures are of less relevance than the later generation steps.
On the other hand, due to the limited one-directional movement in the rectilinear layout, alignment tends to work better with the radial layout, which offers more flexibility in the radial space\footref{supp}.}\remove{
We support both layouts for analysis; the rectilinear is closer to the theoretical representation, while the radial one makes better use of the visual space.}
\addition{Finally, our method substantially reduces the computational time of \removenew{full}\additionnew{global} vanilla t-SNE ($>9x$ per $t$) while providing more visual clarity of data evolution, which highlights both the efficiency and effectiveness of our approach. More analysis on embedding accuracy, displacement, and alignment losses are in the supplementary materials\footref{supp}.}

\section{Use Cases}
\remove{In the previous section, we evaluated and explored the various components of $TDL$.
Here, we illustrate the potential of $TDL$ for analysis of high-level object types, followed by finer attributes and prompt hierarchies to gain deeper insights into the data evolution process.
Specifically, we explore three prompt sets; the first relates to high-level object types or ImageNet classes, the next is finer human facial features, and the last one relates to different styles of cat images.
We use GLIDE~\mbox{\cite{glide}} and Stable Diffusion~\mbox{\cite{stablediff}}
for the first and last two cases, respectively, since they are widely explored and popular open-source text-to-image diffusion models. 
While these two models represent two large classes of underlying diffusion model types, we primarily focus on text-to-image models for our analysis.}
\addition{We demonstrate the potential of \textit{EvolvED} by exploring various research questions, prompts, and diffusion models, to understand the generative process. 
We use both traditional (GLIDE~\cite{glide}) and latent space (Stable Diffusion~\cite{stablediff}) diffusion models. The first introductory case explores simple object evolution with GLIDE.
The latter two explore nuanced feature evolution and prompt hierarchies across different hyperparameter settings using Stable Diffusion.}

\subsection{Exploring ImageNet Objects}
\begin{figure}[t!]
    \centering
    \includegraphics[width=0.99\linewidth]{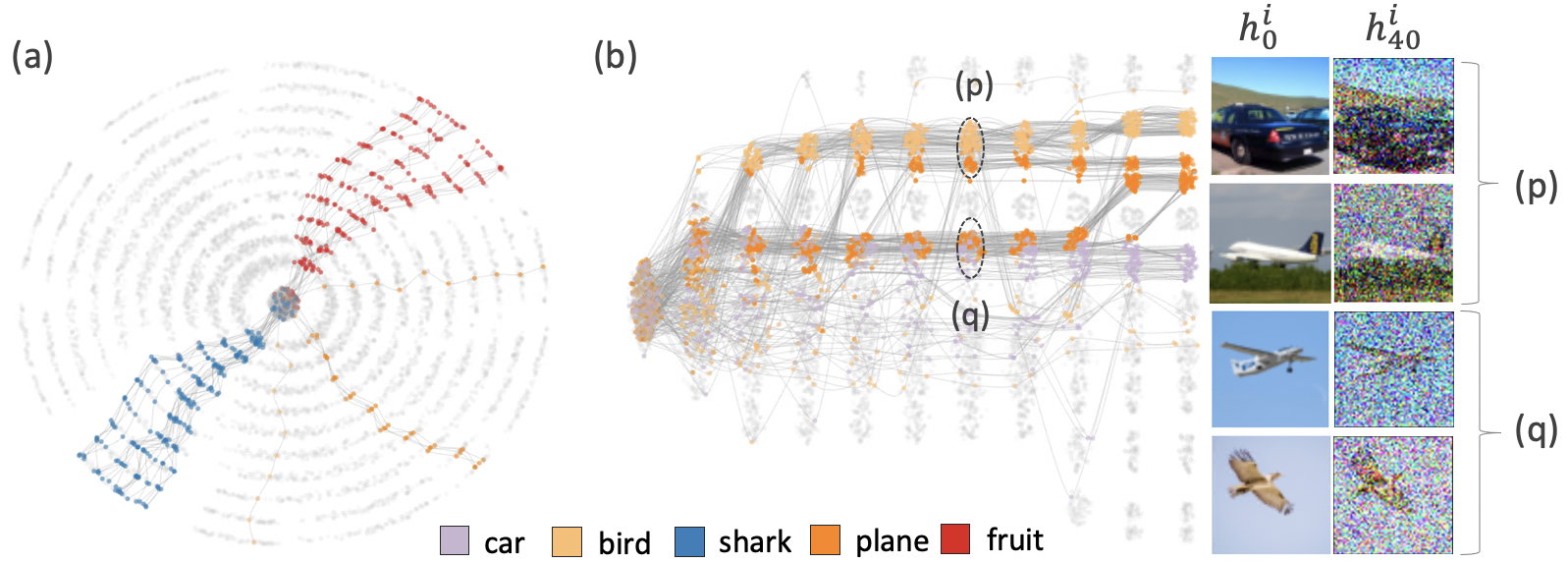}
    \caption{Pathways of instances $\widehat{\textbf{h}}{}^i_{t-1}$ from GLIDE~\cite{glide} encoded with an ImageNet classifier~\cite{bai2021transformers} traced in the evolutionary embeddings:
(a) Pathways shorter than the $5^{th}$ percentile in radial layout, primarily highlighting the $shark$ and $fruit$ instances.
(b) Evolution of $plane$ class and its clusters in the rectilinear layout, with images resembling $bird$ or $car$ at noisy t=40.
\remove{Underwater ($shark$) and land images evolving before the subdivision within land images, i.e., $tiger$s and $dog$s in the (c) radial and (d) rectilinear layouts.} 
}
    \label{fig:imgnetpathways}
\end{figure}

In this section, we study the evolution of distinct high-level object types  using $10$ ImageNet classes, as described in Section~\ref{sec:imagenetdata}.
We start by analyzing the evolutionary embeddings of the noisy instances encoded with the robust classifier~\cite{bai2021transformers} in \cref{fig:imgnetpathways}.
Shorter paths indicate distinct modes, and the longer paths indicate branching and zig-zag patterns.

For instance, the $shark$ and $fruit$ classes exhibit shorter, more direct paths (red and blue points in \cref{fig:imgnetpathways}a), which aligns with their distinct inner circle clusters in \cref{fig:encoders}a due to their distinct colors.
We also observe early differentiation in background features, such as grassy or watery environments, which further subdivide into sub-clusters like $dogs$ and $tigers$ on grass in \cref{fig:encoders} C2 and $sharks$ in water in C4. \remove{This early contextual separation highlights how background context contributes to object distinctions, even in noisy images.}
\begin{figure*}
    \centering
    \includegraphics[width=0.99\textwidth]{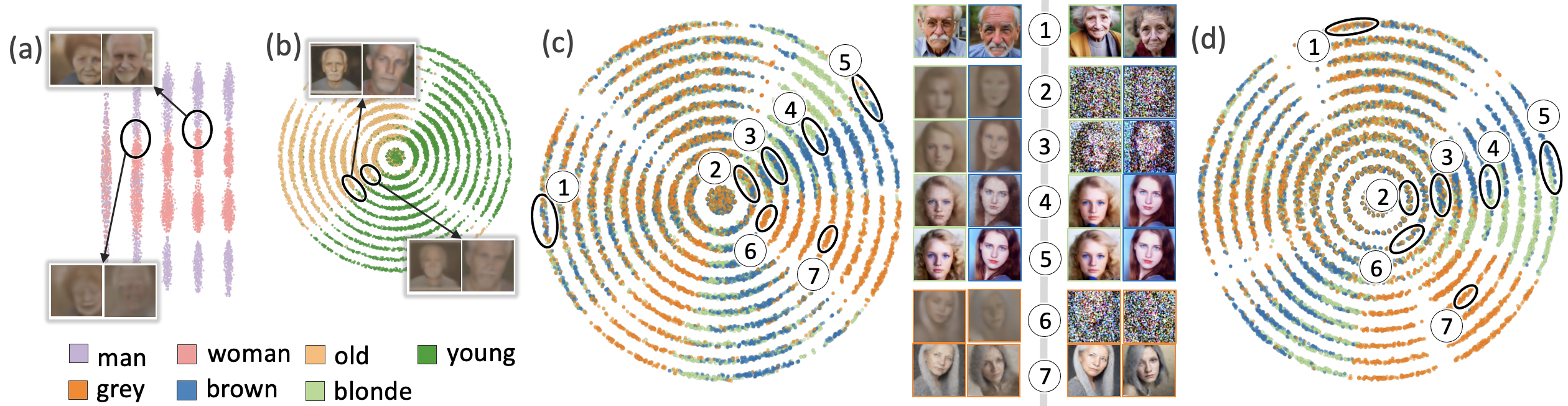}
    \caption{
    Evolution of human facial attributes with Stable Diffusion \addition{and varying schedulers}. 
    \remove{Smooth}\addition{CLIP} representations \addition{of the step-wise denoised images} $\widehat{\textbf{h}}_{t_0}$ are summarized with the evolutionary embedding and color-encoded to show (a) $gender$,\addition{ (b) $age$, and (c) $hair$ $color$, with data generated using a DDPM scheduler.
    The same embedding is generated for the EulerA scheduler.
    The evolution of young $brown$, $blonde$, $grey$ hair (denoised) images is compared between DDPM (C2-C7) and EulerA (D2-D7). 
    Mixed hair clusters for older people are indicated in both scheduler embeddings (C1, D1).
    }\remove{$hair$ $color$ (b) in the radial layout, and $age+gender$ in the radial (c.1) and rectilinear layout (c.2). 
    Sample images at iteration $t$ are highlighted with circles.
    Primary separation is by $gender$ followed by $age$, $grey$ hair, and finally, $blonde$ and $brown$ hair.}
}
    \label{fig:celeb}
\end{figure*}

On exploring the longer paths shown in the rectilinear layout, we identify the $plane$ class, since forming two clusters in the final step (\cref{fig:imgnetpathways}b). Noisy images of planes are split based on the context. They are airborne (seen in the embedding close to birds) or grounded (seen in the embedding close to cars) in \cref{fig:imgnetpathways}q and p. Clear distinctions is observed by the end, the earlier noisy images at $t=40$ are seen as ambiguous, a pattern also noted in the CLIP-encoded features (\cref{fig:encoders}b).
In both cases, the visualization showed that the model seems to resolve context first, followed by finer details, providing an intuitive understanding of findings in literature~\cite{deja2022analyzing, prasad2023unraveling}. 
\remove{Next, we dive deeper into the evolution of finer features to enhance our understanding of the generation process.}

\subsection{Exploring Human Facial Features} 

We explore the hierarchical evolution of features, identifying when \addition{and how }they emerge during the generative process of Stable Diffusion~\cite{stablediff}.
\addition{This could helps assess feature granularity, iteration redundancy, and refine model design. We also explore how hyperparameters, such as schedulers, influence data evolution.}
Our focus is data evolution in the context of human facial features, as this has been explored in literature~\cite{deja2023learning}.\remove{While these works also aimed to reveal such a structure~\mbox{\cite{deja2023learning}}, we aim to provide a more generic approach that is also more interpretable.
We additionally aim to elucidate any entanglement between data attributes relevant for advancing diffusion models~\mbox{\cite{yang2023disdiff}}.}\addition{With this research question, we design prompts that reflect varying levels of granularity in facial attributes.}
Our prompt is defined by the regular expression: $``$a photo of a [$young$ | $old$] [$man$ |  $woman$]; with [$brown$ | $grey$ | $blonde$] hair; with [$brown$ | $blue$] eyes; [\textit{wearing a necklace}]?$"$.
Images are generated in $40$ iterations. 
We further sub-sample $11$ iterations for the embedding and explore the attributes $age$, $gender$, and $hair color$,\remove{ and $eye color$, and whether they were \textit{wearing a necklace}}.
This yielded a total of $48$ prompts representing all possible keyword combinations defined above.
$50$ instances are sampled from each prompt for exploration.
\addition{We choose CLIP to encode attributes of interest, as it has been shown to work with generic facial attributes in previous studies~\cite{agarwal2021evaluatingclip}.}
\newline
\newline
\noindent
\textbf{\addition{Feature evolution:}} \addition{We analyze the progression of facial feature clusters during the diffusion process. Early on, in the third ring, \textit{gender} forms clear clusters, with pink (female) and purple (male) points separating in \cref{fig:celeb}a. By the fourth ring, age-related clusters emerge, with clear separations between younger (light orange) and older (dark green) people in \cref{fig:celeb}b. This is reflected in the denoised image estimates, where age related characteristics emerge over time.
The rectilinear layout in \cref{fig:celeb}a shows clearer early separations due to space but struggles with complex relations because of its one-directional similarity limit, whereas the radial layout in \cref{fig:celeb}b works better for this, highlighting the value of joint analysis.

Hair color clusters show a different pattern; they remain mixed for older people, particularly in \cref{fig:celeb}c1, reflecting a feature entanglement, i.e., $old$ individuals with $brown$ or $blonde$ hair often blend into $grey$. For younger people, distinct hair color clusters appear halfway through the process ($c4$ and $c7$), though some mixing between green and blue points occurs for males in the bottom quartile of \cref{fig:celeb}, likely due to subjective differences in dark blonde vs. brunette or hair volume.
Further, $grey$ hair clusters isolate faster than $blonde$ or $brown$, as shown by the distinct separation in $c6$ versus the mixed grouping in $c3$. This is mirrored in the denoised images, where $brown$ hair in earlier steps resembles a darker blonde.

This feature progression demonstrates the diffusion model's ability to capture low-frequency features like gender early on and later focus on higher-frequency details, a finding consistent with literature~\cite{deja2023learning, prasad2023unraveling}. 
$EvolveD$ provides an intuitive way to pinpoint when specific features emerge and how they compare to others, a capability that is difficult to achieve with existing methods due to their lack of coherence and limited ability to explore trends across iterations.
\newline
\newline
\noindent
\textbf{Exploring schedulers}: The evolution of features in diffusion models depends significantly on the model, hyperparameters, and schedulers used. Schedulers control how noise is added and removed during generation, influencing image quality, diversity, and speed. Understanding their impact is a critical area of research~\cite{duan2023schedulerpath}.
We compare two popular schedulers: DDPM (Denoising Diffusion Probabilistic Model) and Euler Ancestral Discrete Sampling (EulerA). DDPM removes noise in small, linear steps, while EulerA takes larger steps initially but converges faster, leading to more diverse and higher-quality outputs. Visualizing the step-by-step evolution of hair color clusters with $EvolvED$, sheds light on how schedulers manage image-clarity and diversity trade-offs (see \cref{fig:celeb}c and d).

The DDPM embedding shows gradually emerging clusters  (see \cref{fig:celeb} c2 evolving into c3, c4, and c5), which mirrors the gradual changes in the denoised images (see \cref{fig:celeb}c2-c4). In contrast, EulerA embedding displays chaotic, mixed clusters for the first 40\% of steps, as evident from the disordered colors up to the fourth ring (see \cref{fig:celeb}d). We observe that while DDPM gradually refines clusters, EulerA begins with disordered clusters that only start to separate around the fourth ring.
This chaotic behavior in EulerA is reflected in the unstructured, poorly denoised images during early generative steps (see \cref{fig:celeb}d2, d3), indicating the model's limited knowledge of the final image at that stage.
However, this flexibility likely allows EulerA to generate more diverse outputs. 
After this early stage, EulerA converges quickly, producing clearer clusters and sharper, vibrant images in fewer steps in \cref{fig:celeb}d4 compared to DDPM in c4. 
This chaotic start and quick convergence in the embedding overview clearly represents the detailed image-level changes, highlighting the usefulness of $EvolvED$ to study data evolution holistically. 

This analysis, enabled by $EvolvED$, pinpoints key moments when features emerge and refine, offering valuable insights into how each scheduler manages the clarity-diversity tradeoff. While current work on schedulers~\cite{meng2023distillation} look at aggregate noise-level changes, it lacks direct exploration of their impact on high-dimensional data. $EvolvED$ fills this gap by visualizing feature evolution, providing practical insights into how schedulers influence outputs beyond noise scales. Tracking feature evolution helps identify critical refinement stages, which could guide scheduler design or distillation~\cite{meng2023distillation}.}
\subsection{Exploring Image Styles}

\begin{figure}
    \centering
    \includegraphics[width=0.99\linewidth]{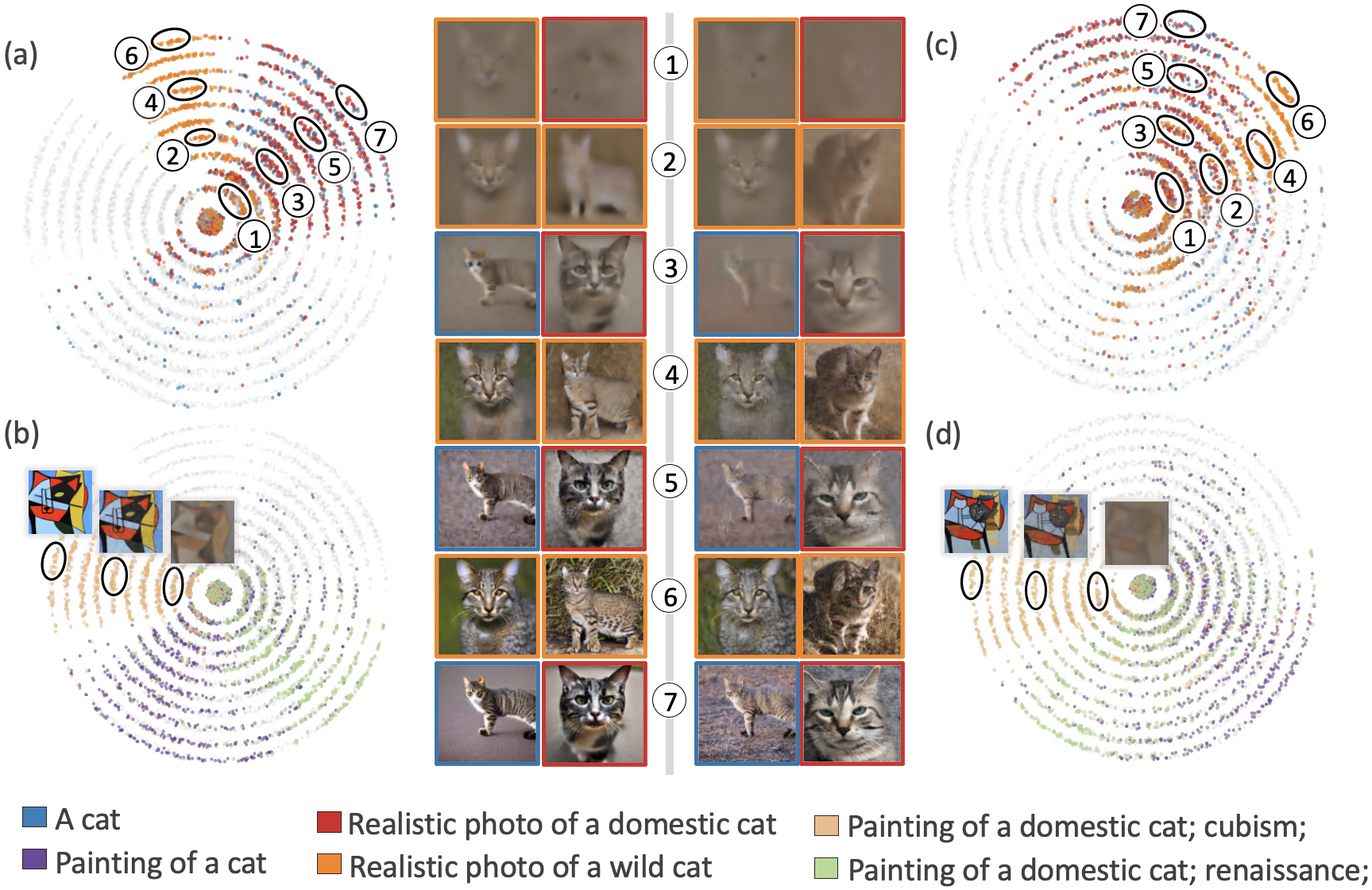}
    \caption{
    Effect of prompt specificity \addition{and guidance scale} on data evolution in StableDiffusion. \addition{Scale is $7.5$ in (a), (b) and $3$ in (c), (d).}
    The evolutionary embedding is based on \remove{smooth}\addition{denoised} representations\remove{$\widehat{\textbf{h}{}^i_{t_0}}$ of images}\remove{, with points color-encoded.
     Images generated with different prompt hierarchies, such as $``$a cat$"$ (a), $``$realistic photo of a cat$"$  (b), and $``$realistic photo of a domestic cat$"$ (c), are clustered in the embedding.
     However, images with the prompt $``$realistic photo of a wild cat$"$ initially mix with others (e) before segregating into separate clusters (d). $``$Painting$"$s of cats (f, g) encompass varying painting styles.}
     \addition{(a) The evolution of $``$a cat$"$ and $``$a realistic photo of a domestic cat$"$, which follow similar paths, while $``$wild cat$"$ images form distinct clusters.
     (b) Painting styles like cubism and renaissance diverge early. Reducing guidance scale in (c), (d) lead to more variability in domestic cat images, now exhibiting wild features, while cubism and wild cat pathways remain stable.}
}
    \label{fig:cats}
\end{figure}

\addition{We shift our research question to explore how hierarchical prompts influence the generative process and pathways within Stable Diffusion~\cite{stablediff}.
Specifically, do generic prompts lead to broader evolutions that encompass those of specific prompts.
We investigate the evolution of (cat) image styles, explored in literature~\cite{lee2023diffexplainer}.
Our goal is to analyze how prompt specificity influences generative pathways, shedding light on learned distributions, potential entanglements, and guiding data generation strategies.
For this purpose, we tailor our prompts as,}
\remove{In this section, we simulate a realistic usage of prompt engineering for image generation by designing prompt hierarchies, going from generic prompts to more specific ones, and examining their impact on the evolutionary generative process in Stable Diffusion~\mbox{\cite{stablediff}}. 
Specifically, we explore cat image styles, which are similar to cases that have also been explored in literature~\mbox{\cite{lee2023diffexplainer}}.
The prompts are defined by the expression} 
$``$A [\textit{realistic photo} of a | \textit{painting} of a]? [\textit{domestic} | \textit{wild}]? cat \remove{[with a \textit{yellow collar} | with an \textit{orange collar}]?} [in the \addition{$cubism$} style\remove{ of \textit{monet}} | in the \addition{$renaissance$} style\remove{ of \textit{van gogh}}]?$"$.
We sampled \remove{$50$}\addition{$100$} instances from each combination. 
\remove{By organizing the prompts in a hierarchical structure, we start with general prompts and gradually introduce more specific ones. 
This approach allows for a more controlled exploration of the styles of cat images, mimicking the exploration of options during the image generation process.
Similar to the previous case, }We apply CLIP on the \remove{smooth representations}\addition{denoised estimates} $\textbf{h}_{t_0}$, which were then summarized through our evolutionary embedding method.
\addition{CLIP was selected for its applicability in style transfer tasks~\cite{patashnik2021styleclip} and generic object-encoding capability, inline with our focus. }
\remove{Our initial exploration centers around analyzing the hierarchy in the specificity of prompts, aiming to understand how certain prompts encompass a larger set of instances compared to others.}
\newline
\newline
\noindent
\textbf{Prompt Specificity \& Generation Pathways:} \Cref{fig:cats}a shows an interesting pattern: blue points representing the prompt $``$a cat$"$ follow almost identical evolution paths as red points representing $``$a realistic photo of a domestic cat$"$\addition{. The images also indicate that both prompts predominantly generate similar domestic cat (denoised) images (see \mbox{\ref{fig:cats}}a1, a3, a5, a7).}\remove{ (see \mbox{\ref{fig:cats_vaguedetail}a, \mbox{\ref{fig:cats_vaguedetail}}b, and \mbox{\ref{fig:cats_vaguedetail}}c)}. 
The corresponding images on the right side of \mbox{\ref{fig:cats_vaguedetail}} indicate that the prompt $``$a cat$"$ primarily generates real images of domestic cats.
While there is some initial mixing between these two sets in cluster~\mbox{\ref{fig:cats_vaguedetail}}e in early iterations until the fifth ring, they subsequently separate into different directions.}
\addition{In contrast, $wild$ cats evolve along a distinct path, with clear clusters in \mbox{\ref{fig:cats}} a2 and a4. 
While initially mixed with domestic cats (see \cref{fig:cats}a1), wild cat features like pointy ears and stretched faces begin to emerge (see \mbox{\cref{fig:cats}a2)} in the denoised images, 30\% into the generation process. 
These features set them apart from the softer features of domestic cat (denoised) images in \mbox{\ref{fig:cats}}a3.}
Analyzing this encompassing aspect of prompts allows us to understand their generated distributions.
In our case, this suggests a skew of cat images toward domestic cats, potentially reflecting the composition of its training data.
\remove{In this case of $``$a cat$"$, our findings suggest a potential bias towards domestic cats, which may have implications depending on the specific use case.}
On the other hand, $painting$s can be seen to evolve differently (see \cref{fig:cats}b) from the $photo$s, as expected.
\addition{Notably, cubism-style paintings take on a unique trajectory, separating early in the generative process (around 20\% into generation), likely due to their bold color patterns. 
This early separation shown in the embedding suggests that specific prompts, such as artistic styles, can guide the generation process toward more distinct outputs, which could be important when aiming for highly specific image characteristics.}
\remove{For example, the generic prompt $``$a painting of a domestic cat$"$ in purple encompasses life-like paintings related to the style of $Monet$ in cluster~\mbox{\ref{fig:cats_vaguedetail}}g, as well as paintings characterized by bold colors related to the style of $van$ $Gogh$ in cluster~\mbox{\ref{fig:cats_vaguedetail}}f.
 Interestingly, $painting$s of cats display a wider variety compared to $photo$s, prompting further exploration of prompt hierarchies within this category (see \mbox{\ref{fig:cats_heirarchy2}}).
Generic prompts, such as $``$a painting of a cat$"$ and $``$a painting of a domestic cat$"$, evolve similarly into those with vibrant colors (see bottom-right of the radial layout in \mbox{\ref{fig:cats_heirarchy2}}).
This finding indicates that the model, by default, generates these more colorful images unless a specific style is specified. 
Conversely, when explicitly specifying a style, such as $monet$, leads to the emergence of softer-toned images in the top-right of the radial layout of \mbox{\ref{fig:cats_heirarchy2}}.}
\newline
\newline
\textbf{Impact of guidance scale}: 
Next, we analyze the guidance scale, as prompt specificity alone may not control image diversity. 
The guidance scale influences how closely the model follows the text prompt. 
By adjusting it, we can explore with $EvolveD$ how generation pathways shift and whether this leads to more flexible or precise outputs, such as combining the domestic and wild cat pathways better.

In \cref{fig:cats}c and \cref{fig:cats}d, we lowered the guidance scale from 7.5 to 3 to observe its impact on the generative process. With less guidance, $EvolvED$'s data evolution visualization helps reveal widened variability from the get go in both the embedding and in the denoised images of the prompts $``$a cat$"$ and $``$a domestic cat$"$. For instance, in \cref{fig:cats}c7, what were previously more controlled domestic cat-like images (see \cref{fig:cats}a7) now exhibit features like nature backgrounds and textured coats, indicating more diverse outputs, reflected by the partial pathway overlap between $domestic$ and $wild$ cats.
Despite this increased variability, the embedding shows that domestic and wild cat prompts follow (less) distinct pathways (see \cref{fig:cats}c4-7).
It also revealed persistent underlying entanglements between photos of cats and domestic cats (\cref{fig:cats}c3-7). This deep entanglement becomes clear through the full process visualization provided by $EvolvED$.
Further, the effect of reducing the guidance scale varies across prompts. For domestic cats, clusters emerge more slowly, as seen in the more fuzzy denoised images in \cref{fig:cats}c2, c4, and c6, compared to earlier formations in \cref{fig:cats}a. However, wild cats (\cref{fig:cats}a2-a6 vs. c2-c6) and cubism paintings are less affected by the reduced guidance (\cref{fig:cats}b vs. d), maintaining stable generative pathways.
This variability in prompt responses to guidance adjustments was made evident through $EvolvED$’s ability to track feature progression over time. 

Generic prompts revealed distribution biases (domestics cat vs. cat), and did not fully encompass specific prompts. The differences in how guidance scales affect the output may stem from the internal model specificity of certain prompts, which retain consistent generative modes, explaining why cubism paintings show less variability. 
 $EvolvED$ was useful in visualizing how prompt specificity and guidance levels influence generation, offering insights beyond what static outputs can reveal. This understanding could inform improvements in training distributions, model adjustments, and better control over generative pathways.

\section{Discussions}
\noindent
\textbf{\remove{TDL}\addition{EvolvED}:} 
 \remove{Concerning $TDL$'s analysis pipeline, the role of an image encoder is essential to study the evolution of specific attributes of data.
Data can evolve in different dimensions of change, for example, shapes, colors, or more generic semantic changes of the images themselves.
Our exploration has predominantly involved simple encoders such as classifiers and foundational image encoding models like CLIP~\mbox{\cite{clip}}.}
While our focus was the evolution of higher-level semantic features like \textit{hair color} in intermediate diffusion images, other evolution dimensions can also be explored, such as U-Net model features~\cite{deja2023learning}, or concept-keyword associations~\cite{park2024timeconcepts}.
Exploring the U-Net features is not straightforward since they do not have a unified latent layer, owing to their skip connections, and different layers being important in different iterations~\cite{si2023freeu, prasad2023unraveling}.
Next, developing advanced visual and interactive design could enhance the analysis of the evolutionary embedding layout generated by \remove{$TDL$}\addition{$EvolvED$}. 
For example, better edge bundling methods could allow the evolution of modes to be less cluttered. 
Images visualized on the outer portions of the layouts could be sampled to be better representatives of the respective cluster. 
Additionally, techniques like word cloud embedding could support the exploration of complex prompt sets.
\newline
\newline
\noindent
\textbf{Evolutionary Embedding:}
We implement the evolutionary embedding with vanilla t-SNE~\cite{van2008tsne} as a proof of concept. 
\remove{However, any dimensionality reduction method that groups semantically similar elements in a low-dimensional space can be utilized.
For example, }\addition{While faster t-SNE~\cite{pezzotti2016approximated, pezzotti2019gpgpu} or UMAP~\cite{umap} should work conceptually since we only add loss functions, we leave this to future work.}
We empirically identified a set of weights the three losses for each layout that performed consistently well across our cases.
However, their weighting balance needs to be considered based on new case requirements.\remove{
Additionally, determining the standard deviation, $\sigma$, for the Gaussian used in the displacement losses ($C{}^c_d$, $C{}^p_d$) depends on user preferences. A larger $\sigma$ is sufficient if clear distinctions between iterations are not crucial, while smaller values may be preferred otherwise.}
\addition{For our alignment constraint, we explore a pairwise loss across consecutive iterations as in prior works~\cite{cantareira2020generic}. However, this can  be improved via a global alignment across iterations.
While rectilinear and radial layouts offer distinct advantages, a formal evaluation of these layouts for representing semantic relations in this context as developed in literature~\cite{pandey2022genorec} may be useful in future work.}
Next, we used Euclidean distances for $C_\text{s}$ in the low-dimensional space of the radial layout.
While empirically, this form was still found to work well, $C_\text{s}$ does not follow the ring structure of the layout and hence can be improved. 
\remove{For example, if two similar points, for example, are initially positioned opposite to each other on a ring, the Euclidean distance metric tends to pull them closer across the rings and not follow the ring structure of the level. 
This is compensated through the displacement loss, $C{}^p_\text{d}$, which is designed to keep points close to the ring.
}A polar-space alternative for $C_\text{s}$ could improve optimization but presents challenges for maintaining gradient correctness.
\newline
\newline
\noindent
\textbf{Scalability:}
Diffusion models consist of several iterations, in the order of $1000$. 
While popular methods~\cite{stablediff, glide} reduce this to $<100$, the size remains substantial for visualization. 
This challenge is particularly observed in radial layouts, where the inner circles are allotted less space.
We sub-sampled iterations in an evenly spaced manner to mitigate visualization difficulties. 
However, more effective interactive sampling methods \addition{or hierarchical dimensionality reduction~\cite{pezzotti2016hierarchical}} can be used. 
\remove{For example, literature~\mbox{\cite{deja2022analyzing,prasad2023unraveling}} indicates a composition phase where content is created, followed by a denoising phase to refine details.
Users could interactively adjust the sampling, potentially selecting more steps initially and fewer later in the process to facilitate more efficient analysis.
Another approach to address visual clutter is to leverage hierarchical embedding methods such as hierarchical stochastic neighborhood embedding or HSNE~\mbox{\cite{pezzotti2016hierarchical}. }}
Finally, the evolution of only a few attributes can be studied at a single time due to limitations in the number of colors used to encode data attributes like \textit{hair color} discernible by the human eye. 
While we reduce this variability with the research question, there is a need for sequential prioritization and analysis of attributes in the current setting.
\newline
\newline
\noindent
\textbf{Future Applications:}
While the main focus of \remove{$TDL$}\addition{$EvolvED$} is to support an understanding of data evolution of diffusion models, its applicability extends to several practically-oriented downstream tasks.
First, exploring branching patterns and when dataset modes emerge are relevant for designing architectures,  noise schedulers, and distillation methods~\cite{karras2022elucidating, duan2023schedulerpath, meng2023distillation}. 
If all modes of interest manifest in early iterations, the number of iterations could be reduced\remove{, or these steps can possibly be skipped entirely}\cite{deja2022analyzing}.
\remove{Next, the exploration of attribute entanglements is widely studied in literature~\mbox{\cite{yang2023disdiff, wu2023uncovering}}.
}\remove{$TDL$}\addition{$EvolvED$} serves as a valuable tool for visualizing attribute entanglements~\cite{yang2023disdiff, wu2023uncovering} and an evolution level. 
Extensions of \remove{$TDL$}\addition{$EvolvED$} could support interactive data generation processes to help users understand and control the generation process.
Lastly, the proposed evolutionary embedding can be extended to diffusion models beyond text-to-image, general DL models feature evolution, model training evolution, and potentially other (non-image) high-dimensional evolutionary data. 


\section{Conclusions}

Diffusion models iteratively reconstruct corrupted data, iteratively evolving noisy images into refined outputs. 
Understanding this data evolution is essential for model understanding and development, but remains complex due to its high-dimensional and iterative nature.
\addition{Existing methods focus on single-step or instance analyses, and do not look at the holistic and iterative aspects together.}\remove{Existing methods like t-SNE to study high-dimensional data do not preserve this iterative structure, limiting its analysis.}
Hence we propose the \remove{\textit{Tree of Diffusion Life} ($TDL$}\addition{$EvolvED$}, a method for holistically understanding data evolution in the generative process of diffusion models. 
\remove{$TDL$}\addition{$EvolvED$} samples the generative space by extracting instances across focused prompts and employs feature extractors to \remove{extract semantic meaning from them}\addition{isolate specific attributes from intermediate images, preserving the iterative context}. 
An evolutionary embedding method is introduced to explicitly encode iterations context while maintaining the \remove{high-dimensional data structure to facilitate the analysis of data evolution. }\addition{feature relations and their evolution to enable analysis.}
Our embedding method achieves this by three main goals: grouping semantically similar elements, organizing elements by iteration, and aligning an instance's elements across iterations.
\remove{We propose rectilinear and radial embeddings to facilitate analysis.
To achieve meaningful semantic distances, images are projected to an intermediate space using image encoders such as classifiers of foundational models like CLIP.
}\addition{This embedding allows us to visualize data evolution in rectilinear and radial layouts, offering a view of the model’s progression over time.}
With different research questions and prompts, we show the versatility and value of \remove{$TDL$}\addition{$EvolvED$} by applying it to two prominent text-to-image diffusion models, GLIDE and StableDiffusion. 
\remove{Looking ahead, $TDL$ sets the basis for potential extension to various practical downstream tasks, such as specific tools to explore entanglements between data attributes or supporting interactive generation.
In summary, $TDL$ shows promise in enhancing our understanding of data evolution in diffusion models and its broader implications.}\addition{\textit{EvolvED} serves as a powerful tool for analyzing data evolution in diffusion models, with applications in understanding feature progression, distributions, and potentially guiding their design.}



\bibliographystyle{abbrv-doi-hyperref}

\bibliography{template}

\begin{thebibliography}{10}

\bibitem{agarwal2021evaluatingclip}
S.~Agarwal, G.~Krueger, J.~Clark, A.~Radford, J.~W. Kim, and M.~Brundage.
\newblock Evaluating clip: towards characterization of broader capabilities and downstream implications.
\newblock {\em arXiv preprint arXiv:2108.02818}, 2021.

\bibitem{bai2021transformers}
Y.~Bai, J.~Mei, A.~L. Yuille, and C.~Xie.
\newblock Are transformers more robust than cnns?
\newblock In M.~Ranzato, A.~Beygelzimer, Y.~Dauphin, P.~Liang, and J.~W. Vaughan, eds., {\em Advances in Neural Information Processing Systems}, vol.~34, pp. 26831--26843. Curran Associates, Inc., 2021. \href{https://doi.org/10.48550/arXiv.2111.05464}
{doi: {{%
10\hspace{.1pt}\discretionary{.}{%
}{.}\hspace{.4pt}48550\discretionary{/}{%
}{/}arXiv\hspace{.1pt}\discretionary{.}{%
}{.}\hspace{.4pt}2111\hspace{.1pt}\discretionary{.}{%
}{.}\hspace{.4pt}05464}}}


\bibitem{cantareira2021explainable}
G.~D. Cantareira, R.~F. Mello, and F.~V. Paulovich.
\newblock Explainable adversarial attacks in deep neural networks using activation profiles.
\newblock {\em arXiv preprint arXiv:2103.10229}, 2021. \href{https://doi.org/10.48550/arXiv.2103.10229}
{doi: {{%
10\hspace{.1pt}\discretionary{.}{%
}{.}\hspace{.4pt}48550\discretionary{/}{%
}{/}arXiv\hspace{.1pt}\discretionary{.}{%
}{.}\hspace{.4pt}2103\hspace{.1pt}\discretionary{.}{%
}{.}\hspace{.4pt}10229}}}


\bibitem{cantareira2020generic}
G.~D. Cantareira and F.~V. Paulovich.
\newblock {A Generic Model for Projection Alignment Applied to Neural Network Visualization}.
\newblock In C.~Turkay and K.~Vrotsou, eds., {\em EuroVis Workshop on Visual Analytics (EuroVA)}. The Eurographics Association, 2020. \href{https://doi.org/10.2312/eurova.20201089}
{doi: {{%
10\hspace{.1pt}\discretionary{.}{%
}{.}\hspace{.4pt}2312\discretionary{/}{%
}{/}eurova\hspace{.1pt}\discretionary{.}{%
}{.}\hspace{.4pt}20201089}}}


\bibitem{chang2023design}
Z.~Chang, G.~A. Koulieris, and H.~P. Shum.
\newblock On the design fundamentals of diffusion models: A survey.
\newblock {\em arXiv preprint arXiv:2306.04542}, 2023. \href{https://doi.org/10.48550/arXiv.2306.04542}
{doi: {{%
10\hspace{.1pt}\discretionary{.}{%
}{.}\hspace{.4pt}48550\discretionary{/}{%
}{/}arXiv\hspace{.1pt}\discretionary{.}{%
}{.}\hspace{.4pt}2306\hspace{.1pt}\discretionary{.}{%
}{.}\hspace{.4pt}04542}}}


\bibitem{chefer2023attend}
H.~Chefer, Y.~Alaluf, Y.~Vinker, L.~Wolf, and D.~Cohen-Or.
\newblock Attend-and-excite: Attention-based semantic guidance for text-to-image diffusion models.
\newblock {\em ACM Transactions on Graphics}, 42(4),  article no. 148,  10 pages, jul 2023. \href{https://doi.org/10.1145/3592116}
{doi: {{%
10\hspace{.1pt}\discretionary{.}{%
}{.}\hspace{.4pt}1145\discretionary{/}{%
}{/}3592116}}}


\bibitem{deja2022analyzing}
K.~Deja, A.~Kuzina, T.~Trzcinski, and J.~Tomczak.
\newblock On analyzing generative and denoising capabilities of diffusion-based deep generative models.
\newblock In S.~Koyejo, S.~Mohamed, A.~Agarwal, D.~Belgrave, K.~Cho, and A.~Oh, eds., {\em Advances in Neural Information Processing Systems}, vol.~35, pp. 26218--26229. Curran Associates, Inc., 2022. \href{https://doi.org/10.48550/arXiv.2206.00070}
{doi: {{%
10\hspace{.1pt}\discretionary{.}{%
}{.}\hspace{.4pt}48550\discretionary{/}{%
}{/}arXiv\hspace{.1pt}\discretionary{.}{%
}{.}\hspace{.4pt}2206\hspace{.1pt}\discretionary{.}{%
}{.}\hspace{.4pt}00070}}}


\bibitem{deja2023learning}
K.~Deja, T.~Trzci{\'{n}}ski, and J.~M. Tomczak.
\newblock Learning data representations with joint diffusion models.
\newblock In D.~Koutra, C.~Plant, M.~Gomez~Rodriguez, E.~Baralis, and F.~Bonchi, eds., {\em Machine Learning and Knowledge Discovery in Databases: Research Track}, pp. 543--559. Springer Nature Switzerland, Cham, 2023. \href{https://doi.org/10.1007/978-3-031-43415-0_32}
{doi: {{%
10\hspace{.1pt}\discretionary{.}{%
}{.}\hspace{.4pt}1007\discretionary{/}{%
}{/}978\discretionary{%
}{-}{-}3\discretionary{%
}{-}{-}031\discretionary{%
}{-}{-}43415\discretionary{%
}{-}{-}0\_32}}}


\bibitem{duan2023schedulerpath}
Z.~Duan, C.~Wang, C.~Chen, J.~Huang, and W.~Qian.
\newblock Optimal linear subspace search: Learning to construct fast and high-quality schedulers for diffusion models.
\newblock In {\em Proceedings of the 32nd ACM International Conference on Information and Knowledge Management}, pp. 463--472, 2023.

\bibitem{fang2020survey}
Y.~Fang, H.~Xu, and J.~Jiang.
\newblock A survey of time series data visualization research.
\newblock {\em IOP Conference Series: Materials Science and Engineering}, 782(2):022013, mar 2020. \href{https://doi.org/10.1088/1757-899X/782/2/022013}
{doi: {{%
10\hspace{.1pt}\discretionary{.}{%
}{.}\hspace{.4pt}1088\discretionary{/}{%
}{/}1757\discretionary{%
}{-}{-}899X\discretionary{/}{%
}{/}782\discretionary{/}{%
}{/}2\discretionary{/}{%
}{/}022013}}}


\bibitem{hertz2022prompt}
A.~Hertz, R.~Mokady, J.~Tenenbaum, K.~Aberman, Y.~Pritch, and D.~Cohen-Or.
\newblock Prompt-to-prompt image editing with cross attention control.
\newblock {\em arXiv preprint arXiv:2208.01626}, 2022. \href{https://doi.org/10.48550/arXiv.2208.01626}
{doi: {{%
10\hspace{.1pt}\discretionary{.}{%
}{.}\hspace{.4pt}48550\discretionary{/}{%
}{/}arXiv\hspace{.1pt}\discretionary{.}{%
}{.}\hspace{.4pt}2208\hspace{.1pt}\discretionary{.}{%
}{.}\hspace{.4pt}01626}}}


\bibitem{hinterreiter2021projection}
A.~Hinterreiter, C.~Steinparz, M.~Sch\"{O}fl, H.~Stitz, and M.~Streit.
\newblock Projection path explorer: Exploring visual patterns in projected decision-making paths.
\newblock {\em ACM Trans. Interact. Intell. Syst.}, 11(3–4),  article no. 22,  29 pages, sep 2021. \href{https://doi.org/10.1145/3387165}
{doi: {{%
10\hspace{.1pt}\discretionary{.}{%
}{.}\hspace{.4pt}1145\discretionary{/}{%
}{/}3387165}}}


\bibitem{ho2020denoising}
J.~Ho, A.~Jain, and P.~Abbeel.
\newblock Denoising diffusion probabilistic models.
\newblock In H.~Larochelle, M.~Ranzato, R.~Hadsell, M.~Balcan, and H.~Lin, eds., {\em Advances in Neural Information Processing Systems}, vol.~33, pp. 6840--6851. Curran Associates, 2020. \href{https://doi.org/10.48550/arXiv.2006.11239}
{doi: {{%
10\hspace{.1pt}\discretionary{.}{%
}{.}\hspace{.4pt}48550\discretionary{/}{%
}{/}arXiv\hspace{.1pt}\discretionary{.}{%
}{.}\hspace{.4pt}2006\hspace{.1pt}\discretionary{.}{%
}{.}\hspace{.4pt}11239}}}


\bibitem{kahng2018gan}
M.~Kahng, N.~Thorat, D.~H. Chau, F.~B. Viégas, and M.~Wattenberg.
\newblock Gan lab: Understanding complex deep generative models using interactive visual experimentation.
\newblock {\em IEEE Transactions on Visualization and Computer Graphics}, 25(1):310--320, 2019. \href{https://doi.org/10.1109/TVCG.2018.2864500}
{doi: {{%
10\hspace{.1pt}\discretionary{.}{%
}{.}\hspace{.4pt}1109\discretionary{/}{%
}{/}TVCG\hspace{.1pt}\discretionary{.}{%
}{.}\hspace{.4pt}2018\hspace{.1pt}\discretionary{.}{%
}{.}\hspace{.4pt}2864500}}}


\bibitem{karras2022elucidating}
T.~Karras, M.~Aittala, T.~Aila, and S.~Laine.
\newblock Elucidating the design space of diffusion-based generative models.
\newblock {\em Advances in Neural Information Processing Systems}, 35:26565--26577, 2022. \href{https://doi.org/10.48550/arXiv.2206.00364}
{doi: {{%
10\hspace{.1pt}\discretionary{.}{%
}{.}\hspace{.4pt}48550\discretionary{/}{%
}{/}arXiv\hspace{.1pt}\discretionary{.}{%
}{.}\hspace{.4pt}2206\hspace{.1pt}\discretionary{.}{%
}{.}\hspace{.4pt}00364}}}


\bibitem{kaski2003trustworthiness}
S.~Kaski, J.~Nikkil{\"a}, M.~Oja, J.~Venna, P.~T{\"o}r{\"o}nen, and E.~Castr{\'e}n.
\newblock Trustworthiness and metrics in visualizing similarity of gene expression.
\newblock {\em BMC bioinformatics}, 4:1--13, 2003. \href{https://doi.org/10.1186/1471-2105-4-48}
{doi: {{%
10\hspace{.1pt}\discretionary{.}{%
}{.}\hspace{.4pt}1186\discretionary{/}{%
}{/}1471\discretionary{%
}{-}{-}2105\discretionary{%
}{-}{-}4\discretionary{%
}{-}{-}48}}}


\bibitem{kwon2023diffusion}
M.~Kwon, J.~Jeong, and Y.~Uh.
\newblock Diffusion models already have a semantic latent space.
\newblock In {\em The Eleventh International Conference on Learning Representations}, 2023. \href{https://doi.org/10.48550/arXiv.2210.10960}
{doi: {{%
10\hspace{.1pt}\discretionary{.}{%
}{.}\hspace{.4pt}48550\discretionary{/}{%
}{/}arXiv\hspace{.1pt}\discretionary{.}{%
}{.}\hspace{.4pt}2210\hspace{.1pt}\discretionary{.}{%
}{.}\hspace{.4pt}10960}}}


\bibitem{lee2023diffexplainer}
S.~Lee, B.~Hoover, H.~Strobelt, Z.~J. Wang, S.~Peng, A.~Wright, K.~Li, H.~Park, H.~Yang, and D.~H. Chau.
\newblock Diffusion explainer: Visual explanation for text-to-image stable diffusion.
\newblock {\em arXiv preprint arXiv:2305.03509}, 2023. \href{https://doi.org/10.48550/arXiv.2305.03509}
{doi: {{%
10\hspace{.1pt}\discretionary{.}{%
}{.}\hspace{.4pt}48550\discretionary{/}{%
}{/}arXiv\hspace{.1pt}\discretionary{.}{%
}{.}\hspace{.4pt}2305\hspace{.1pt}\discretionary{.}{%
}{.}\hspace{.4pt}03509}}}


\bibitem{li2020visualizing}
M.~Li, Z.~Zhao, and C.~Scheidegger.
\newblock Visualizing neural networks with the grand tour.
\newblock {\em Distill}, 5(3):e25, 2020.

\bibitem{luccioni2023stablebias}
S.~Luccioni, C.~Akiki, M.~Mitchell, and Y.~Jernite.
\newblock Stable bias: Evaluating societal representations in diffusion models.
\newblock In {\em Thirty-seventh Conference on Neural Information Processing Systems Datasets and Benchmarks Track}, 2023. \href{https://doi.org/10.48550/arXiv.2303.11408}
{doi: {{%
10\hspace{.1pt}\discretionary{.}{%
}{.}\hspace{.4pt}48550\discretionary{/}{%
}{/}arXiv\hspace{.1pt}\discretionary{.}{%
}{.}\hspace{.4pt}2303\hspace{.1pt}\discretionary{.}{%
}{.}\hspace{.4pt}11408}}}


\bibitem{luo2022dimenfix}
Q.~Luo, L.~Christino, F.~V. Paulovich, and E.~Milios.
\newblock Dimenfix: A novel meta-dimensionality reduction method for feature preservation.
\newblock {\em arXiv preprint arXiv:2211.16752}, 2022. \href{https://doi.org/10.48550/arXiv.2211.16752}
{doi: {{%
10\hspace{.1pt}\discretionary{.}{%
}{.}\hspace{.4pt}48550\discretionary{/}{%
}{/}arXiv\hspace{.1pt}\discretionary{.}{%
}{.}\hspace{.4pt}2211\hspace{.1pt}\discretionary{.}{%
}{.}\hspace{.4pt}16752}}}


\bibitem{umap}
L.~McInnes, J.~Healy, N.~Saul, and L.~Großberger.
\newblock Umap: Uniform manifold approximation and projection.
\newblock {\em Journal of Open Source Software}, 3(29):861, 2018. \href{https://doi.org/10.21105/joss.00861}
{doi: {{%
10\hspace{.1pt}\discretionary{.}{%
}{.}\hspace{.4pt}21105\discretionary{/}{%
}{/}joss\hspace{.1pt}\discretionary{.}{%
}{.}\hspace{.4pt}00861}}}


\bibitem{meng2023distillation}
C.~Meng, R.~Rombach, R.~Gao, D.~Kingma, S.~Ermon, J.~Ho, and T.~Salimans.
\newblock On distillation of guided diffusion models.
\newblock In {\em Proceedings of the IEEE/CVF Conference on Computer Vision and Pattern Recognition}, pp. 14297--14306, 2023.

\bibitem{meng2023class}
L.~Meng, S.~van~den Elzen, N.~Pezzotti, and A.~Vilanova.
\newblock Class-constrained t-sne: Combining data features and class probabilities.
\newblock {\em IEEE Transactions on Visualization and Computer Graphics}, 30(1):164--174, 2024. \href{https://doi.org/10.1109/TVCG.2023.3326600}
{doi: {{%
10\hspace{.1pt}\discretionary{.}{%
}{.}\hspace{.4pt}1109\discretionary{/}{%
}{/}TVCG\hspace{.1pt}\discretionary{.}{%
}{.}\hspace{.4pt}2023\hspace{.1pt}\discretionary{.}{%
}{.}\hspace{.4pt}3326600}}}


\bibitem{nichol2021improved}
A.~Q. Nichol and P.~Dhariwal.
\newblock Improved denoising diffusion probabilistic models.
\newblock In {\em International Conference on Machine Learning}, pp. 8162--8171. PMLR, 2021. \href{https://doi.org/10.48550/arXiv.2102.09672}
{doi: {{%
10\hspace{.1pt}\discretionary{.}{%
}{.}\hspace{.4pt}48550\discretionary{/}{%
}{/}arXiv\hspace{.1pt}\discretionary{.}{%
}{.}\hspace{.4pt}2102\hspace{.1pt}\discretionary{.}{%
}{.}\hspace{.4pt}09672}}}


\bibitem{glide}
A.~Q. Nichol, P.~Dhariwal, A.~Ramesh, P.~Shyam, P.~Mishkin, B.~Mcgrew, I.~Sutskever, and M.~Chen.
\newblock {GLIDE}: Towards photorealistic image generation and editing with text-guided diffusion models.
\newblock In {\em Proceedings of the 39th International Conference on Machine Learning}, Proceedings of Machine Learning Research. PMLR, 2022. \href{https://doi.org/10.48550/arXiv.2112.10741}
{doi: {{%
10\hspace{.1pt}\discretionary{.}{%
}{.}\hspace{.4pt}48550\discretionary{/}{%
}{/}arXiv\hspace{.1pt}\discretionary{.}{%
}{.}\hspace{.4pt}2112\hspace{.1pt}\discretionary{.}{%
}{.}\hspace{.4pt}10741}}}


\bibitem{pandey2022genorec}
A.~Pandey, S.~L'Yi, Q.~Wang, M.~A. Borkin, and N.~Gehlenborg.
\newblock Genorec: a recommendation system for interactive genomics data visualization.
\newblock {\em IEEE transactions on visualization and computer graphics}, 29(1):570--580, 2022.

\bibitem{park2024timeconcepts}
J.-H. Park, Y.-J. Ju, and S.-W. Lee.
\newblock Explaining generative diffusion models via visual analysis for interpretable decision-making process.
\newblock {\em Expert Systems with Applications}, 248:123231, 2024. \href{https://doi.org/10.1016/j.eswa.2024.123231}
{doi: {{%
10\hspace{.1pt}\discretionary{.}{%
}{.}\hspace{.4pt}1016\discretionary{/}{%
}{/}j\hspace{.1pt}\discretionary{.}{%
}{.}\hspace{.4pt}eswa\hspace{.1pt}\discretionary{.}{%
}{.}\hspace{.4pt}2024\hspace{.1pt}\discretionary{.}{%
}{.}\hspace{.4pt}123231}}}


\bibitem{park2024riemannian}
Y.-H. Park, M.~Kwon, J.~Choi, J.~Jo, and Y.~Uh.
\newblock Understanding the latent space of diffusion models through the lens of riemannian geometry.
\newblock In A.~Oh, T.~Neumann, A.~Globerson, K.~Saenko, M.~Hardt, and S.~Levine, eds., {\em Advances in Neural Information Processing Systems}, vol.~36, pp. 24129--24142. Curran Associates, Inc., 2023. \href{https://doi.org/10.48550/arXiv.2307.12868}
{doi: {{%
10\hspace{.1pt}\discretionary{.}{%
}{.}\hspace{.4pt}48550\discretionary{/}{%
}{/}arXiv\hspace{.1pt}\discretionary{.}{%
}{.}\hspace{.4pt}2307\hspace{.1pt}\discretionary{.}{%
}{.}\hspace{.4pt}12868}}}


\bibitem{parmar2023zero}
G.~Parmar, K.~Kumar~Singh, R.~Zhang, Y.~Li, J.~Lu, and J.-Y. Zhu.
\newblock Zero-shot image-to-image translation.
\newblock In {\em ACM SIGGRAPH 2023 Conference Proceedings}, SIGGRAPH '23,  article no. 11,  11 pages. Association for Computing Machinery, New York, NY, USA, 2023. \href{https://doi.org/10.1145/3588432.3591513}
{doi: {{%
10\hspace{.1pt}\discretionary{.}{%
}{.}\hspace{.4pt}1145\discretionary{/}{%
}{/}3588432\hspace{.1pt}\discretionary{.}{%
}{.}\hspace{.4pt}3591513}}}


\bibitem{patashnik2021styleclip}
O.~Patashnik, Z.~Wu, E.~Shechtman, D.~Cohen-Or, and D.~Lischinski.
\newblock Styleclip: Text-driven manipulation of stylegan imagery.
\newblock In {\em Proceedings of the IEEE/CVF international conference on computer vision}, pp. 2085--2094, 2021.

\bibitem{pezzotti2018multiscale}
N.~Pezzotti, J.-D. Fekete, T.~H{\"o}llt, B.~P. Lelieveldt, E.~Eisemann, and A.~Vilanova.
\newblock Multiscale visualization and exploration of large bipartite graphs.
\newblock In {\em Computer Graphics Forum}, vol.~37, pp. 549--560. Wiley Online Library, 2018. \href{https://doi.org/10.1111/cgf.13441}
{doi: {{%
10\hspace{.1pt}\discretionary{.}{%
}{.}\hspace{.4pt}1111\discretionary{/}{%
}{/}cgf\hspace{.1pt}\discretionary{.}{%
}{.}\hspace{.4pt}13441}}}


\bibitem{pezzotti2016hierarchical}
N.~Pezzotti, T.~H{\"o}llt, B.~Lelieveldt, E.~Eisemann, and A.~Vilanova.
\newblock Hierarchical stochastic neighbor embedding.
\newblock In {\em Computer Graphics Forum}, vol.~35, pp. 21--30. Wiley Online Library, 2016. \href{https://doi.org/10.1111/cgf.12878}
{doi: {{%
10\hspace{.1pt}\discretionary{.}{%
}{.}\hspace{.4pt}1111\discretionary{/}{%
}{/}cgf\hspace{.1pt}\discretionary{.}{%
}{.}\hspace{.4pt}12878}}}


\bibitem{pezzotti2017deepeyes}
N.~Pezzotti, T.~Höllt, J.~Van~Gemert, B.~P. Lelieveldt, E.~Eisemann, and A.~Vilanova.
\newblock Deepeyes: Progressive visual analytics for designing deep neural networks.
\newblock {\em IEEE Transactions on Visualization and Computer Graphics}, 24(1):98--108, 2018. \href{https://doi.org/10.1109/TVCG.2017.2744358}
{doi: {{%
10\hspace{.1pt}\discretionary{.}{%
}{.}\hspace{.4pt}1109\discretionary{/}{%
}{/}TVCG\hspace{.1pt}\discretionary{.}{%
}{.}\hspace{.4pt}2017\hspace{.1pt}\discretionary{.}{%
}{.}\hspace{.4pt}2744358}}}


\bibitem{pezzotti2016approximated}
N.~Pezzotti, B.~P.~F. Lelieveldt, L.~v.~d. Maaten, T.~Höllt, E.~Eisemann, and A.~Vilanova.
\newblock Approximated and user steerable tsne for progressive visual analytics.
\newblock {\em IEEE Transactions on Visualization and Computer Graphics}, 23(7):1739--1752, 2017. \href{https://doi.org/10.1109/TVCG.2016.2570755}
{doi: {{%
10\hspace{.1pt}\discretionary{.}{%
}{.}\hspace{.4pt}1109\discretionary{/}{%
}{/}TVCG\hspace{.1pt}\discretionary{.}{%
}{.}\hspace{.4pt}2016\hspace{.1pt}\discretionary{.}{%
}{.}\hspace{.4pt}2570755}}}


\bibitem{pezzotti2019gpgpu}
N.~Pezzotti, J.~Thijssen, A.~Mordvintsev, T.~Höllt, B.~Van~Lew, B.~P. Lelieveldt, E.~Eisemann, and A.~Vilanova.
\newblock Gpgpu linear complexity t-sne optimization.
\newblock {\em IEEE Transactions on Visualization and Computer Graphics}, 26(1):1172--1181, 2020. \href{https://doi.org/10.1109/TVCG.2019.2934307}
{doi: {{%
10\hspace{.1pt}\discretionary{.}{%
}{.}\hspace{.4pt}1109\discretionary{/}{%
}{/}TVCG\hspace{.1pt}\discretionary{.}{%
}{.}\hspace{.4pt}2019\hspace{.1pt}\discretionary{.}{%
}{.}\hspace{.4pt}2934307}}}


\bibitem{prasad2022transform}
V.~Prasad, R.~J. van Sloun, S.~van~den Elzen, A.~Vilanova, and N.~Pezzotti.
\newblock The transform-and-perform framework: Explainable deep learning beyond classification.
\newblock {\em IEEE Transactions on Visualization and Computer Graphics}, 30(2):1502--1515, 2022. \href{https://doi.org/10.1109/TVCG.2022.3219248}
{doi: {{%
10\hspace{.1pt}\discretionary{.}{%
}{.}\hspace{.4pt}1109\discretionary{/}{%
}{/}TVCG\hspace{.1pt}\discretionary{.}{%
}{.}\hspace{.4pt}2022\hspace{.1pt}\discretionary{.}{%
}{.}\hspace{.4pt}3219248}}}


\bibitem{prasad2023unraveling}
V.~Prasad, C.~Zhu-Tian, A.~Vilanova, H.~Pfister, N.~Pezzotti, and H.~Strobelt.
\newblock Unraveling the temporal dynamics of the unet in diffusion models, 2024. \href{https://doi.org/10.48550/arXiv.2312.14965}
{doi: {{%
10\hspace{.1pt}\discretionary{.}{%
}{.}\hspace{.4pt}48550\discretionary{/}{%
}{/}arXiv\hspace{.1pt}\discretionary{.}{%
}{.}\hspace{.4pt}2312\hspace{.1pt}\discretionary{.}{%
}{.}\hspace{.4pt}14965}}}


\bibitem{clip}
A.~Radford, J.~W. Kim, C.~Hallacy, A.~Ramesh, G.~Goh, S.~Agarwal, G.~Sastry, A.~Askell, P.~Mishkin, J.~Clark, et~al.
\newblock Learning transferable visual models from natural language supervision.
\newblock In {\em International conference on machine learning}, pp. 8748--8763. PMLR, 2021. \href{https://doi.org/10.48550/arXiv.2103.00020}
{doi: {{%
10\hspace{.1pt}\discretionary{.}{%
}{.}\hspace{.4pt}48550\discretionary{/}{%
}{/}arXiv\hspace{.1pt}\discretionary{.}{%
}{.}\hspace{.4pt}2103\hspace{.1pt}\discretionary{.}{%
}{.}\hspace{.4pt}00020}}}


\bibitem{rauber2016visualizing}
P.~E. Rauber, S.~G. Fadel, A.~X. Falcão, and A.~C. Telea.
\newblock Visualizing the hidden activity of artificial neural networks.
\newblock {\em IEEE Transactions on Visualization and Computer Graphics}, 23(1):101--110, 2017. \href{https://doi.org/10.1109/TVCG.2016.2598838}
{doi: {{%
10\hspace{.1pt}\discretionary{.}{%
}{.}\hspace{.4pt}1109\discretionary{/}{%
}{/}TVCG\hspace{.1pt}\discretionary{.}{%
}{.}\hspace{.4pt}2016\hspace{.1pt}\discretionary{.}{%
}{.}\hspace{.4pt}2598838}}}


\bibitem{stablediff}
R.~Rombach, A.~Blattmann, D.~Lorenz, P.~Esser, and B.~Ommer.
\newblock High-resolution image synthesis with latent diffusion models.
\newblock In {\em Proceedings of the IEEE/CVF Conference on Computer Vision and Pattern Recognition (CVPR)}, pp. 10684--10695, June 2022. \href{https://doi.org/10.1109/CVPR52688.2022.01042}
{doi: {{%
10\hspace{.1pt}\discretionary{.}{%
}{.}\hspace{.4pt}1109\discretionary{/}{%
}{/}CVPR52688\hspace{.1pt}\discretionary{.}{%
}{.}\hspace{.4pt}2022\hspace{.1pt}\discretionary{.}{%
}{.}\hspace{.4pt}01042}}}


\bibitem{unet}
O.~Ronneberger, P.~Fischer, and T.~Brox.
\newblock U-net: Convolutional networks for biomedical image segmentation.
\newblock In N.~Navab, J.~Hornegger, W.~M. Wells, and A.~F. Frangi, eds., {\em Medical Image Computing and Computer-Assisted Intervention}, pp. 234--241. Springer, Cham, 2015. \href{https://doi.org/10.1007/978-3-319-24574-4_28}
{doi: {{%
10\hspace{.1pt}\discretionary{.}{%
}{.}\hspace{.4pt}1007\discretionary{/}{%
}{/}978\discretionary{%
}{-}{-}3\discretionary{%
}{-}{-}319\discretionary{%
}{-}{-}24574\discretionary{%
}{-}{-}4\_28}}}


\bibitem{si2023freeu}
C.~Si, Z.~Huang, Y.~Jiang, and Z.~Liu.
\newblock Freeu: Free lunch in diffusion u-net.
\newblock {\em arXiv preprint arXiv:2309.11497}, 2023. \href{https://doi.org/10.48550/arXiv.2309.11497}
{doi: {{%
10\hspace{.1pt}\discretionary{.}{%
}{.}\hspace{.4pt}48550\discretionary{/}{%
}{/}arXiv\hspace{.1pt}\discretionary{.}{%
}{.}\hspace{.4pt}2309\hspace{.1pt}\discretionary{.}{%
}{.}\hspace{.4pt}11497}}}


\bibitem{sohl2015deep}
J.~Sohl-Dickstein, E.~Weiss, N.~Maheswaranathan, and S.~Ganguli.
\newblock Deep unsupervised learning using nonequilibrium thermodynamics.
\newblock In {\em International conference on machine learning}, pp. 2256--2265. PMLR, 2015. \href{https://doi.org/10.48550/arXiv.1503.03585}
{doi: {{%
10\hspace{.1pt}\discretionary{.}{%
}{.}\hspace{.4pt}48550\discretionary{/}{%
}{/}arXiv\hspace{.1pt}\discretionary{.}{%
}{.}\hspace{.4pt}1503\hspace{.1pt}\discretionary{.}{%
}{.}\hspace{.4pt}03585}}}


\bibitem{song2021scorebased}
Y.~Song, J.~Sohl-Dickstein, D.~P. Kingma, A.~Kumar, S.~Ermon, and B.~Poole.
\newblock Score-based generative modeling through stochastic differential equations.
\newblock In {\em International Conference on Learning Representations}, 2021. \href{https://doi.org/10.48550/arXiv.2011.13456}
{doi: {{%
10\hspace{.1pt}\discretionary{.}{%
}{.}\hspace{.4pt}48550\discretionary{/}{%
}{/}arXiv\hspace{.1pt}\discretionary{.}{%
}{.}\hspace{.4pt}2011\hspace{.1pt}\discretionary{.}{%
}{.}\hspace{.4pt}13456}}}


\bibitem{tang2023daam}
R.~Tang, L.~Liu, A.~Pandey, Z.~Jiang, G.~Yang, K.~Kumar, P.~Stenetorp, J.~Lin, and F.~Ture.
\newblock What the {DAAM}: Interpreting stable diffusion using cross attention.
\newblock In A.~Rogers, J.~Boyd-Graber, and N.~Okazaki, eds., {\em Proceedings of the 61st Annual Meeting of the Association for Computational Linguistics}, pp. 5644--5659. Association for Computational Linguistics, Toronto, Canada, July 2023. \href{https://doi.org/10.18653/v1/2023.acl-long.310}
{doi: {{%
10\hspace{.1pt}\discretionary{.}{%
}{.}\hspace{.4pt}18653\discretionary{/}{%
}{/}v1\discretionary{/}{%
}{/}2023\hspace{.1pt}\discretionary{.}{%
}{.}\hspace{.4pt}acl\discretionary{%
}{-}{-}long\hspace{.1pt}\discretionary{.}{%
}{.}\hspace{.4pt}310}}}


\bibitem{Tumanyan_2023_CVPR}
N.~Tumanyan, M.~Geyer, S.~Bagon, and T.~Dekel.
\newblock Plug-and-play diffusion features for text-driven image-to-image translation.
\newblock In {\em 2023 IEEE/CVF Conference on Computer Vision and Pattern Recognition (CVPR)}, pp. 1921--1930, 2023. \href{https://doi.org/10.1109/CVPR52729.2023.00191}
{doi: {{%
10\hspace{.1pt}\discretionary{.}{%
}{.}\hspace{.4pt}1109\discretionary{/}{%
}{/}CVPR52729\hspace{.1pt}\discretionary{.}{%
}{.}\hspace{.4pt}2023\hspace{.1pt}\discretionary{.}{%
}{.}\hspace{.4pt}00191}}}


\bibitem{vahdat2021score}
A.~Vahdat, K.~Kreis, and J.~Kautz.
\newblock Score-based generative modeling in latent space.
\newblock {\em Advances in neural information processing systems}, 34:11287--11302, 2021. \href{https://doi.org/10.48550/arXiv.2106.05931}
{doi: {{%
10\hspace{.1pt}\discretionary{.}{%
}{.}\hspace{.4pt}48550\discretionary{/}{%
}{/}arXiv\hspace{.1pt}\discretionary{.}{%
}{.}\hspace{.4pt}2106\hspace{.1pt}\discretionary{.}{%
}{.}\hspace{.4pt}05931}}}


\bibitem{van2008tsne}
L.~van~der Maaten and G.~Hinton.
\newblock Visualizing data using t-sne.
\newblock {\em Journal of Machine Learning Research}, 9(86):2579--2605, 2008.

\bibitem{videau2023interactive}
M.~Videau, N.~Knizev, A.~Leite, M.~Schoenauer, and O.~Teytaud.
\newblock Interactive latent diffusion model.
\newblock In {\em Proceedings of the Genetic and Evolutionary Computation Conference}, GECCO '23,  11 pages, p. 586–596. Association for Computing Machinery, New York, NY, USA, 2023. \href{https://doi.org/10.1145/3583131.3590471}
{doi: {{%
10\hspace{.1pt}\discretionary{.}{%
}{.}\hspace{.4pt}1145\discretionary{/}{%
}{/}3583131\hspace{.1pt}\discretionary{.}{%
}{.}\hspace{.4pt}3590471}}}


\bibitem{wang2020cnn}
Z.~J. Wang, R.~Turko, O.~Shaikh, H.~Park, N.~Das, F.~Hohman, M.~Kahng, and D.~H. Polo~Chau.
\newblock Cnn explainer: Learning convolutional neural networks with interactive visualization.
\newblock {\em IEEE Transactions on Visualization and Computer Graphics}, 27(2):1396--1406, 2021. \href{https://doi.org/10.1109/TVCG.2020.3030418}
{doi: {{%
10\hspace{.1pt}\discretionary{.}{%
}{.}\hspace{.4pt}1109\discretionary{/}{%
}{/}TVCG\hspace{.1pt}\discretionary{.}{%
}{.}\hspace{.4pt}2020\hspace{.1pt}\discretionary{.}{%
}{.}\hspace{.4pt}3030418}}}


\bibitem{wu2023uncovering}
Q.~Wu, Y.~Liu, H.~Zhao, A.~Kale, T.~Bui, T.~Yu, Z.~Lin, Y.~Zhang, and S.~Chang.
\newblock Uncovering the disentanglement capability in text-to-image diffusion models.
\newblock In {\em 2023 IEEE/CVF Conference on Computer Vision and Pattern Recognition}, pp. 1900--1910. IEEE Computer Society, Los Alamitos, CA, USA, jun 2023. \href{https://doi.org/10.1109/CVPR52729.2023.00189}
{doi: {{%
10\hspace{.1pt}\discretionary{.}{%
}{.}\hspace{.4pt}1109\discretionary{/}{%
}{/}CVPR52729\hspace{.1pt}\discretionary{.}{%
}{.}\hspace{.4pt}2023\hspace{.1pt}\discretionary{.}{%
}{.}\hspace{.4pt}00189}}}


\bibitem{yang2023diffusionsurvey}
L.~Yang, Z.~Zhang, Y.~Song, S.~Hong, R.~Xu, Y.~Zhao, W.~Zhang, B.~Cui, and M.-H. Yang.
\newblock Diffusion models: A comprehensive survey of methods and applications.
\newblock {\em ACM Computing Surveys}, 56(4),  article no. 105,  39 pages, nov 2023. \href{https://doi.org/10.1145/3626235}
{doi: {{%
10\hspace{.1pt}\discretionary{.}{%
}{.}\hspace{.4pt}1145\discretionary{/}{%
}{/}3626235}}}


\bibitem{yang2023disdiff}
T.~Yang, Y.~Wang, Y.~Lu, and N.~Zheng.
\newblock Disdiff: Unsupervised disentanglement of diffusion probabilistic models.
\newblock In {\em Thirty-seventh Conference on Neural Information Processing Systems}, 2023. \href{https://doi.org/10.48550/arXiv.2301.13721}
{doi: {{%
10\hspace{.1pt}\discretionary{.}{%
}{.}\hspace{.4pt}48550\discretionary{/}{%
}{/}arXiv\hspace{.1pt}\discretionary{.}{%
}{.}\hspace{.4pt}2301\hspace{.1pt}\discretionary{.}{%
}{.}\hspace{.4pt}13721}}}


\bibitem{zhao2024cupid}
Y.~Zhao, M.~Li, and M.~Berger.
\newblock Cupid: Contextual understanding of prompt-conditioned image distributions.
\newblock In {\em Computer Graphics Forum}, p. e15086. Wiley Online Library, 2024.

\end{thebibliography}
\section*{Supplemental Materials}
\label{sec:supplemental_materials}
\begin{figure*}[!h]
    \centering
    \includegraphics[width=0.95\linewidth]{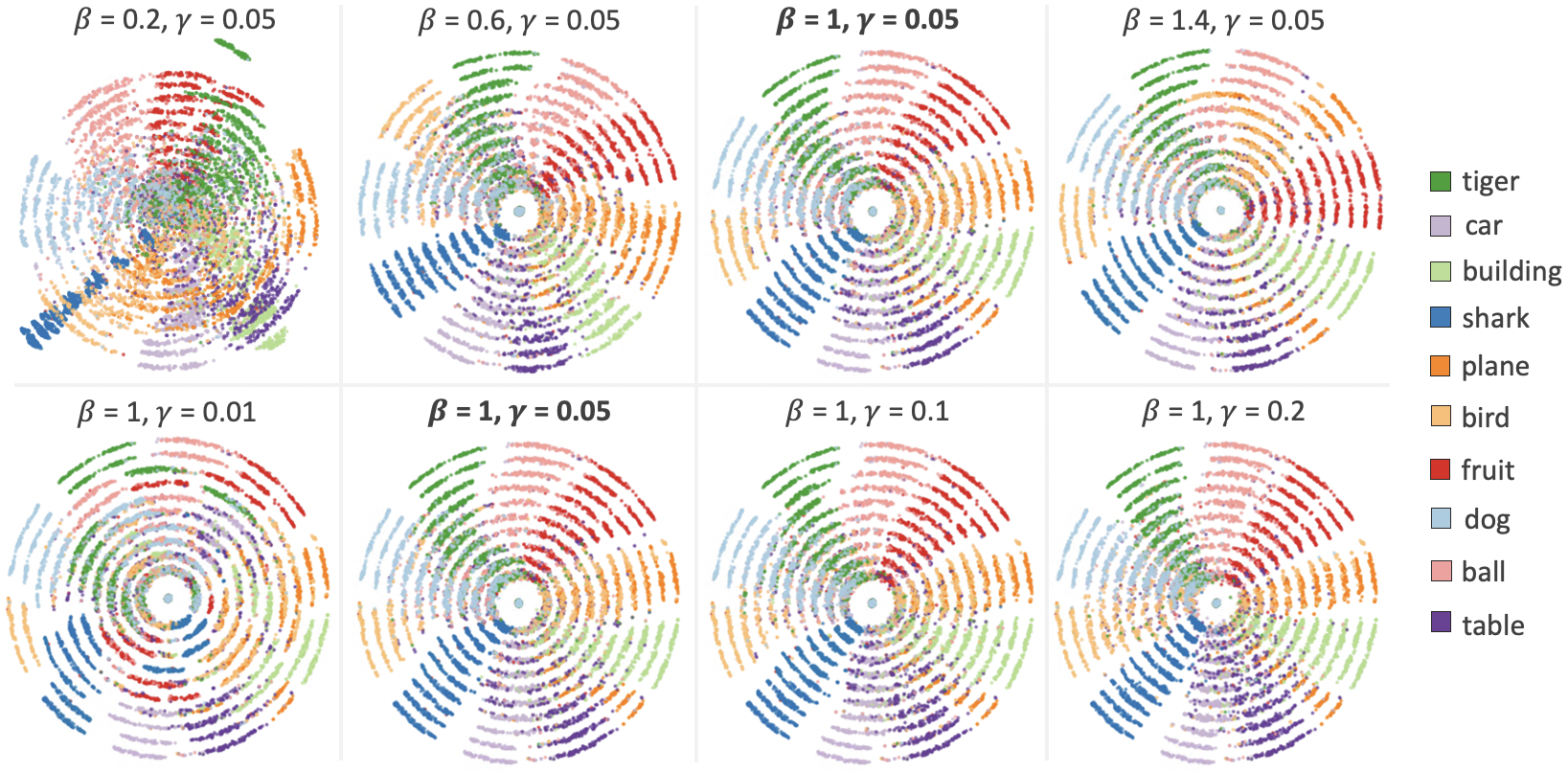}
    \captionof{figure}{
    \textbf{Exploring different $\beta$ and $\gamma$ weightings for the radial layout} losses $C{}^p_d$ and $C{}^p_a$ respectively, with $\alpha=1$.  
    When $\beta=0.2$, the iterations are poorly separated. 
    Increasing $\beta=0.6$ enhances separation, yet inner clusters remain insufficiently distinguished. 
    While $\beta\in\{1, 1.4\}$, the iterations display well-defined separation.
    $\beta=1$ is chosen to maintain the t-SNE cluster structure while meeting our objectives.
    Similarly, with $\gamma=0.01$, points demonstrate poor alignment, hindering the exploration of data evolution. 
    Increasing $\gamma$ values of $0.05$ and $0.1$ leads to sufficient alignment. $\gamma = 0.05$ is selected based on the above insights and neighborhood quality metrics.
    All embeddings in this document are on the ImageNet objects case, where the noisy representations $\widehat{h}{}^i_t$ ($t=99$ to $t=0$) are first encoded with a classifier.
    While the cosine alignment $C_a$ loss is normalized ($[0,1]$), converting it back into Cartesian coordinates for gradient updates causes the step sizes to vary depending on the radius size, as angular distances change in the Euclidean space across radii. To accommodate t-SNE, which optimizes better when points are closer to the center, we use a fixed radius of $\overline{r}_{t}=20$ across iterations for optimization and stretch out the radii for visualization; our chosen $\gamma=0.05$ is based on this. Changing $\overline{r}_{t}$ is expected to influence $\gamma$ and requires careful tuning. Since $\overline{r}_{t}=20$ has been empirically verified to work well with t-SNE, we recommend maintaining this value for optimization. }
    \label{fig:rad_betgam_search}
\end{figure*}

\begin{figure*}[!h]
    \centering
    \includegraphics[width=0.95\linewidth]{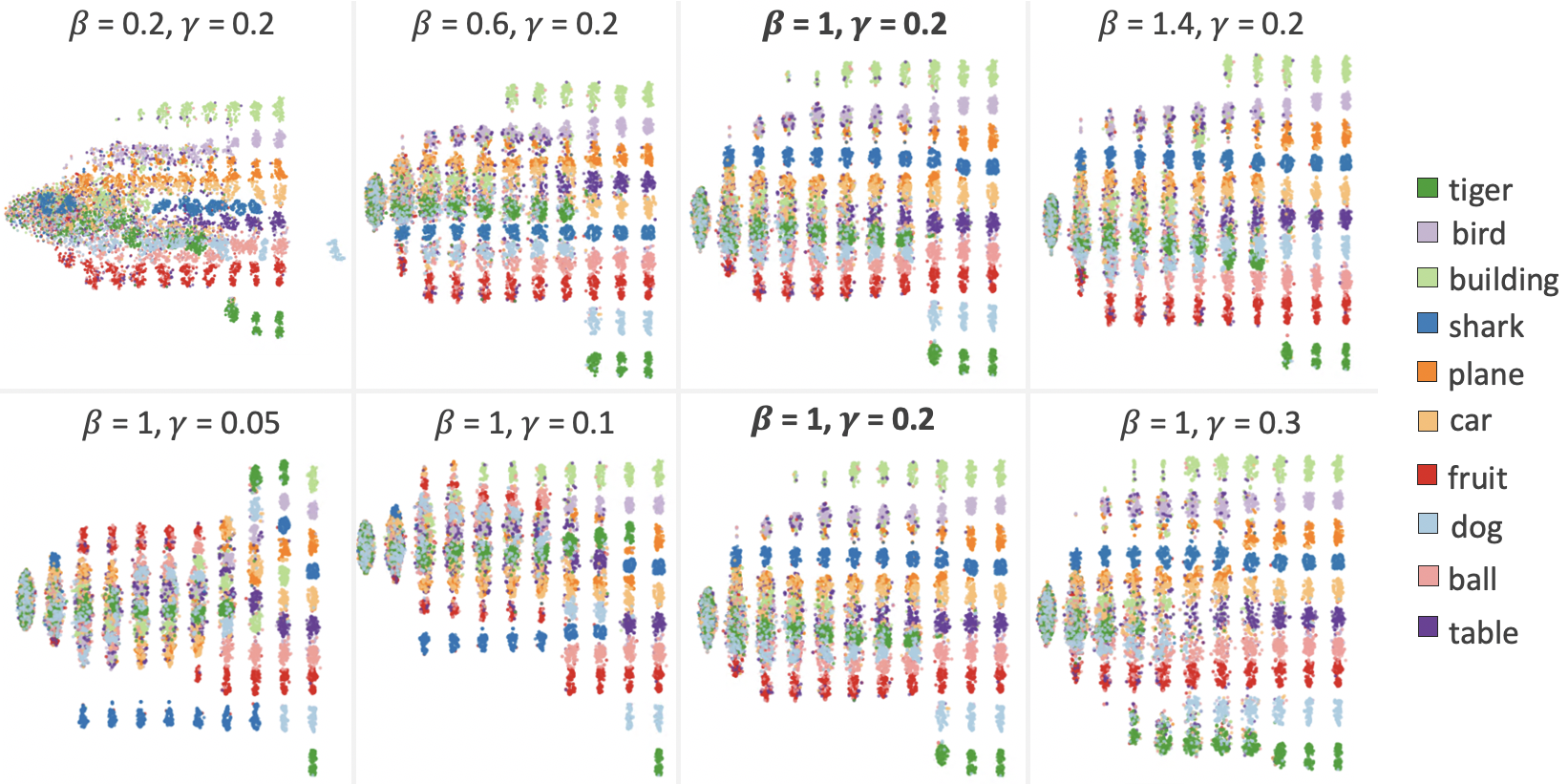}
    \caption{
    \textbf{Exploring different $\beta$ and $\gamma$ weightings for the rectilinear layout} losses $C{}^c_d$ and $C{}^c_a$ respectively, with $\alpha=1$.
    Similar to previous experiments, for $\beta\in\{1,3\}$, the iterations exhibit poor separation, with unclear links between points and iteration steps. 
    A minimum value of $\beta=1$ is necessary to achieve sufficient separation and is hence selected.
    $\gamma=0.05$ and $\gamma=0.1$ result in inadequate alignment of clusters across iterations. 
    While $\gamma=0.2$ and $\gamma=0.3$ produce satisfactory results, we choose $\gamma=0.2$ to minimize disruptions to the t-SNE layout while allowing for alignment. 
    The alignments are not smooth in some areas (see $\beta=1$, $\gamma=0.2$), such as the blue-green points between the $4^{th}$ and $3^{rd}$ lines from the right. Alignment is a tradeoff in these layouts, as limited movement makes optimization challenging. 
    Increasing $\gamma=0.3$ results in smoother alignments, but this comes at the expense of cluster quality, which is not desirable.}
    \label{fig:rect_betgam_search}
\end{figure*}

\begin{figure*}
    \centering
    \includegraphics[width=0.95\linewidth]{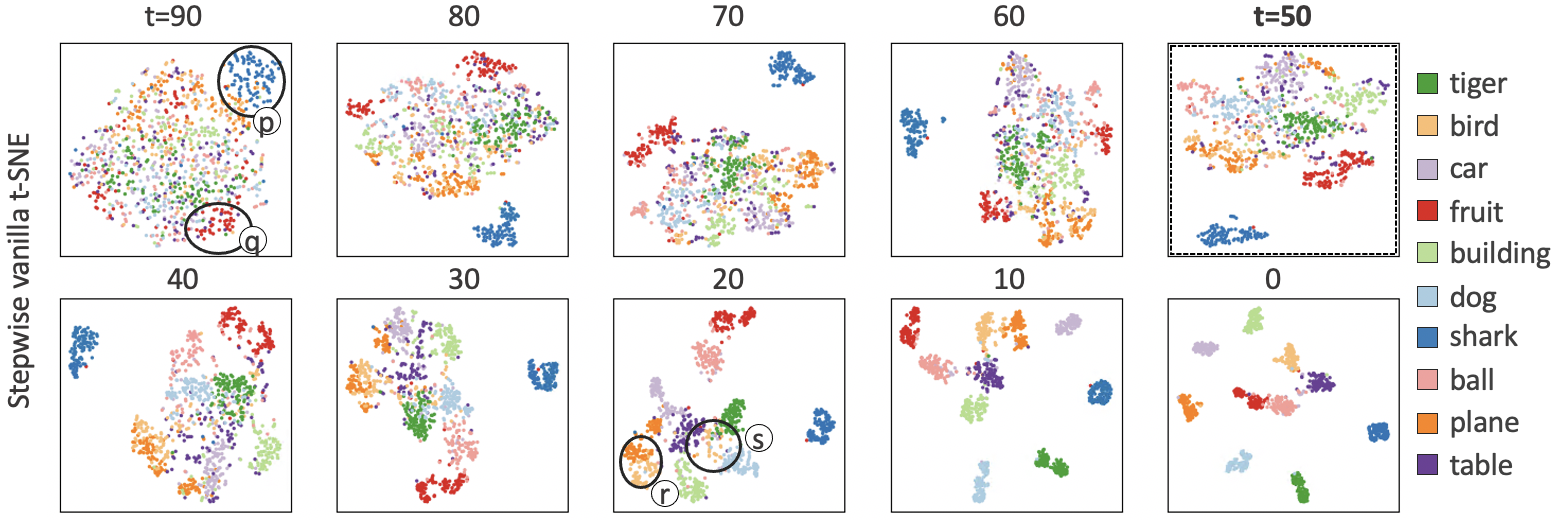}
    \caption{
    \textbf{Visualizing the individual embeddings of the baseline stepwise vanilla t-SNE} for comparison with the proposed layouts.
    This shows the affect of $C_s$ alone, i.e., $\alpha=1$, $\beta=0$, and $\gamma=0$. Note that show embeddings from the second sampled time step (we skip $t=100$). 
    We observe that the overall cluster structures remain similar to the radial ($\alpha=1$, $\beta=1$, $\gamma=0.05$), and rectilinear layouts ($\alpha=1$, $\beta=1$, $\gamma=0.2$).
    For example, in all three the $dark$ $blue$ and $red$ points of sharks and fruits emerge by the third line from the left or inner ring, i.e., $t=80$. Also,
    the $dark$ and $light$ $orange$ points of planes and birds are mixed until $t=30$.
    This reflects the quality of our evolutionary embedding in maintaing clusters, while enabling the study of data progression more easily.}
    \label{fig:stepwisetsne}
\end{figure*}
\begin{figure*}[!h]
    \centering
    \includegraphics[width=0.95\linewidth]{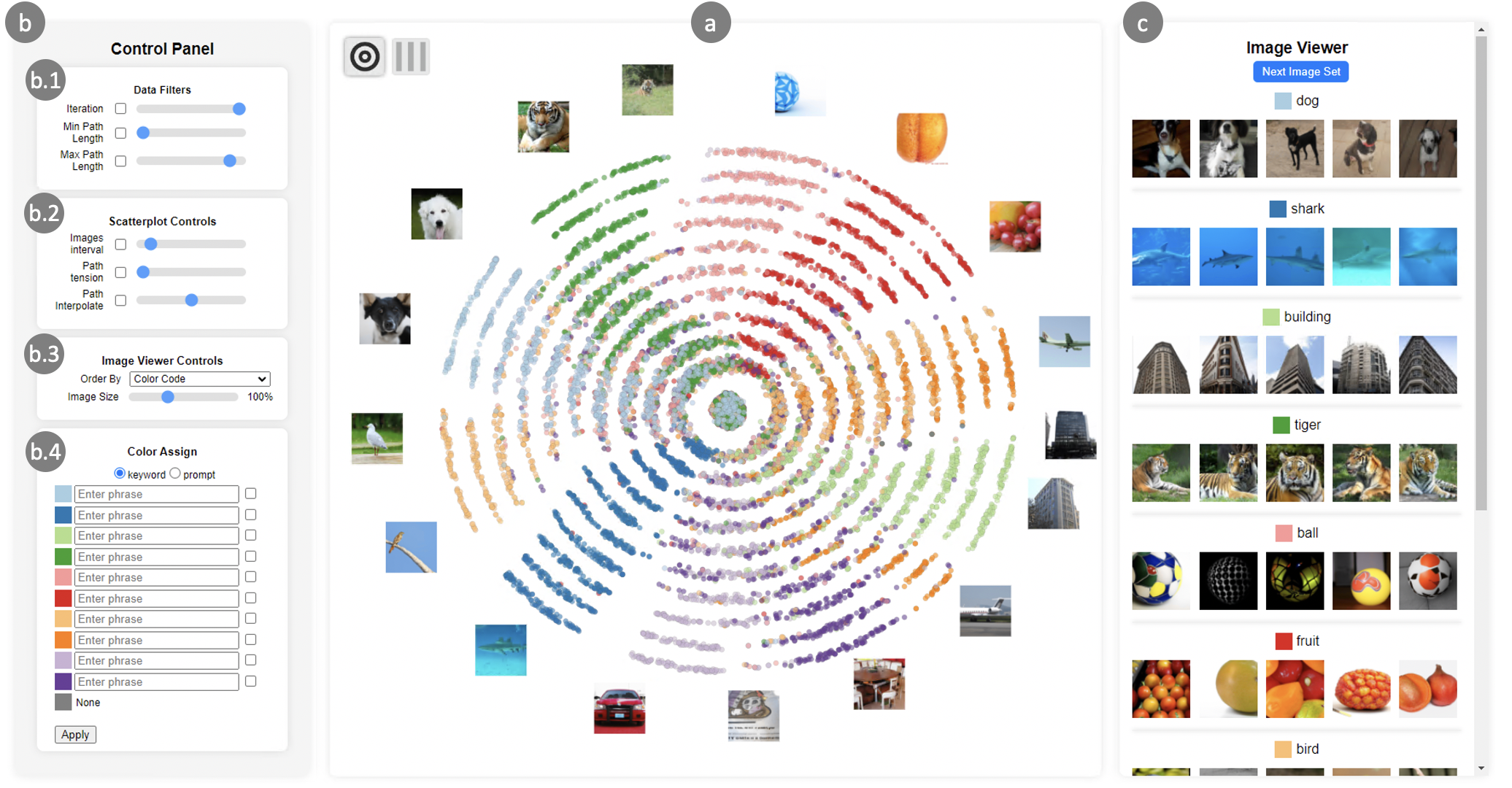}
    \captionof{figure}{
    \textbf{Our proof of concept front-end} enables analysis of the evolutionary embeddings produced by $EvolvED$.
    At the center (a) is our proposed evolutionary embedding, which is the outcome of $EvolvED$. 
    The above shows the radial layout, but the rectilinear one can be retrieved with the buttons on the top-right of (a). 
    Images corresponding to points at every $n$ degree are displayed along the outer ring of the layout (a).
    A control panel on the left (b) supports data filtering, visualizing, and coloring options.
    Selected images are displayed on the right (c) in an image viewer to enable understanding of the embedding.
    Points can be filtered based on the iteration $t$, the path length (b.1), or prompt keywords (b.4).
    Some visualization options, such as deciding the number of images to be displayed on the outer ring, the smoothness of the paths, and the interpolation factor to group paths of a cluster, are provided (b.2).  
    Points in the embedding can be encoded by adjustable colors based on the prompt or a keyword within a prompt (b.4). 
    Lastly, images in the image viewer can be grouped by time step or color code to enable analysis (b.3). 
    }
    \label{fig:frontend}
\end{figure*}

\begin{figure}
    \centering
    \includegraphics[width=0.95\linewidth]{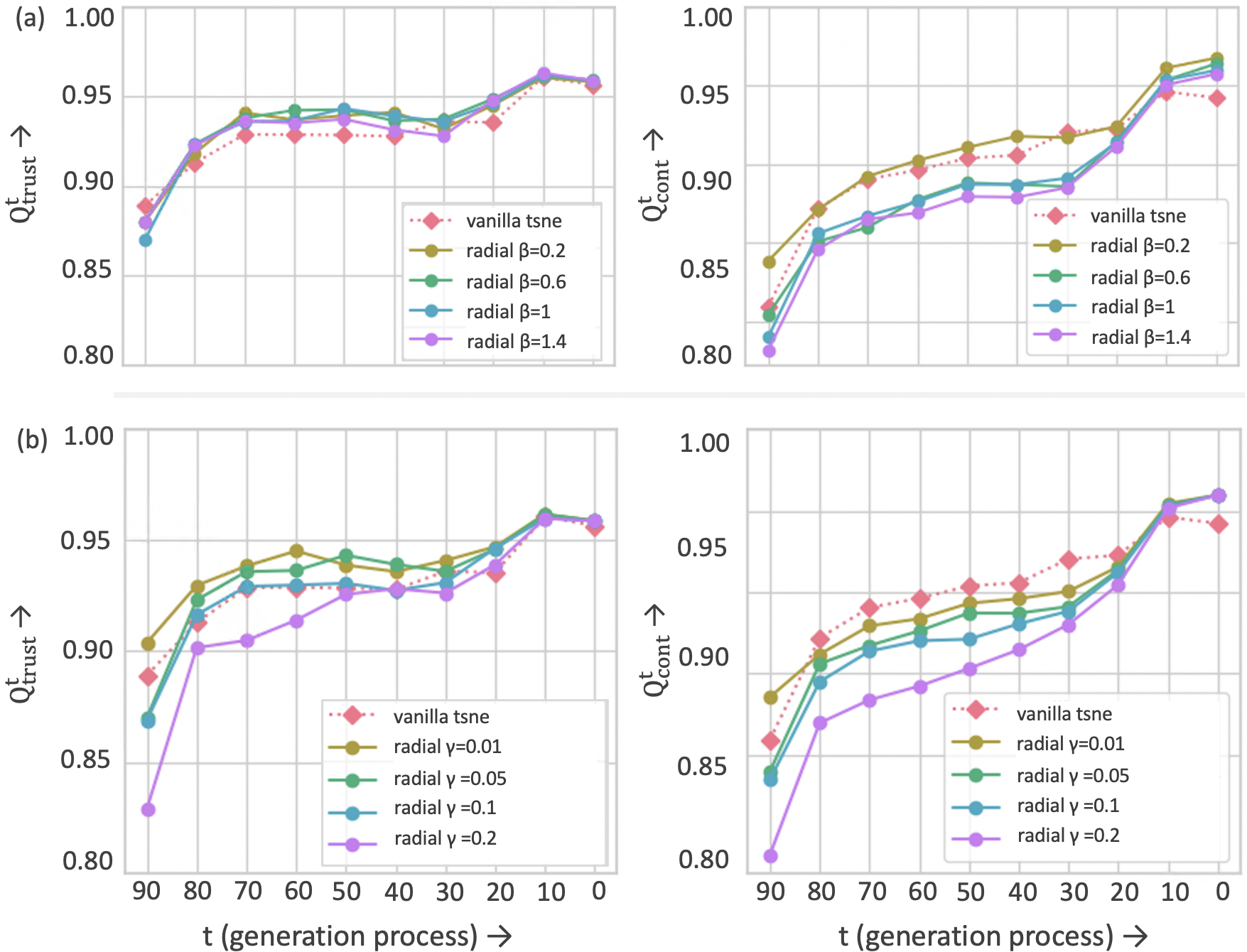}
    \caption{
    \textbf{Evaluating neighborhood preservation with $Q{}^t_\text{trust}$ and $Q{}^t_\text{cont}$ for the proposed radial layout} across (a) different $\beta$ values with $\alpha=1$, $\gamma=0.05$, and (b) different $\gamma$ values with $\alpha=1$, $\beta=1$.
    Embeddings are derived from $\widehat{\textbf{h}}{}^i_{t-1}$.
    While $Q{}^t_\text{trust}$ across various $\beta$ values in (a) appear comparable, $Q{}^t_\text{cont}$ exhibits noticeable variations. 
    Despite $\beta=0.2$ yielding the best $Q{}^t_\text{cont}$ values, it resulted in poor visual separability between iterations shown in Fig.~\ref{fig:rad_betgam_search}. Conversely, $\beta \in {0.6,1}$ are comparable in preserving high-dimensional neighborhoods. 
    Hence, $\beta=1$ was selected for improved iteration separation seen in Fig.~\ref{fig:rad_betgam_search}.
    Further, $Q{}^t_\text{cont}$ and $Q{}^t_\text{trust}$ in (b) display degradation with $\gamma=0.2$, while $\gamma \in {0.01, 0.05}$ remain comparable in preserving neighborhoods. 
    We hence choose $\gamma=0.05$ since lower values result in insufficient alignment.
    }
    \label{fig:rad_quant_metrics}
\end{figure}

\begin{figure}
    \centering
    \includegraphics[width=0.95\linewidth]{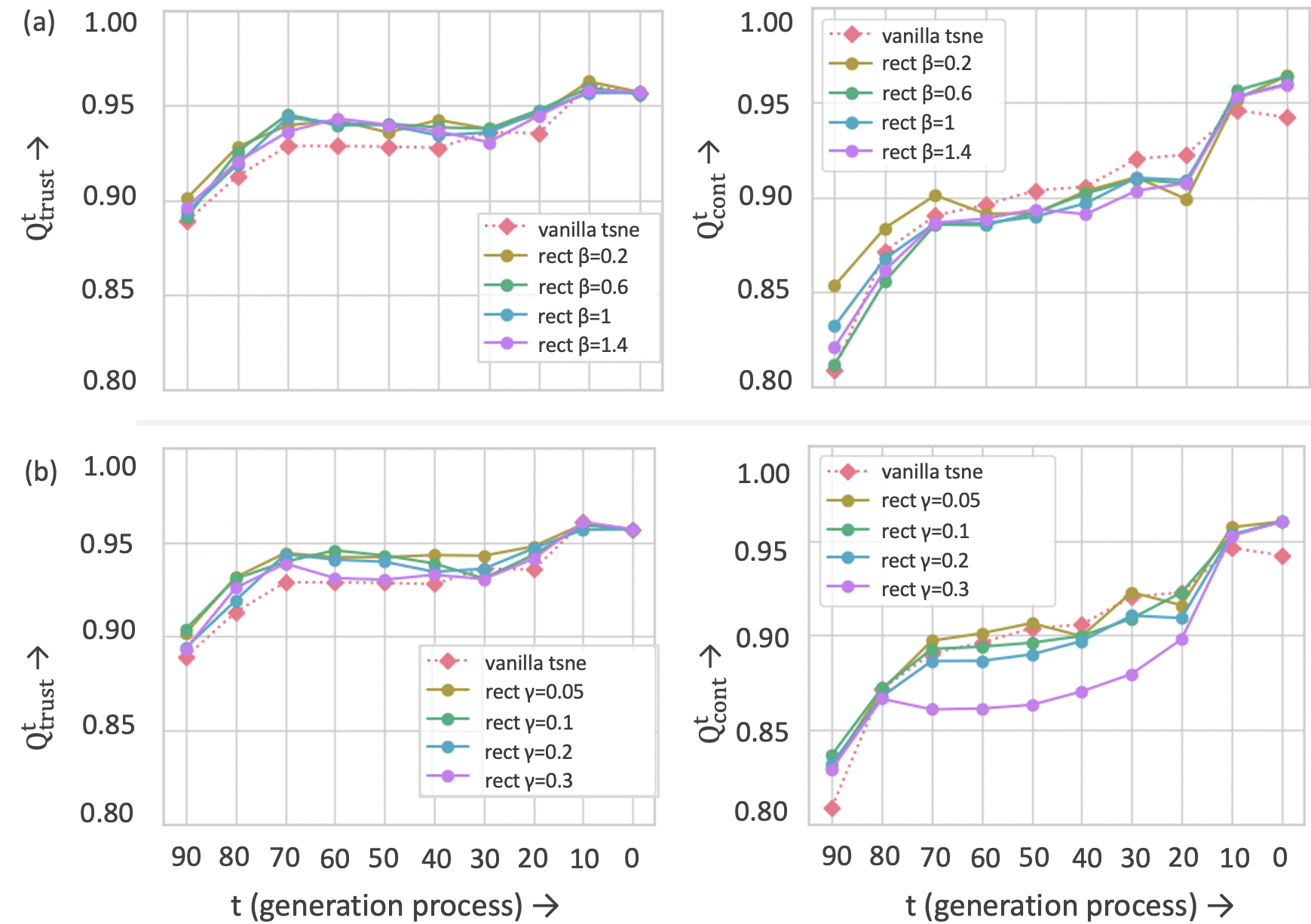}
    \caption{
    \textbf{Evaluating neighborhood preservation with $Q{}^t_\text{trust}$ and $Q{}^t_\text{cont}$ for the proposed rectilinear layout} across (a) different $\beta$ values with $\alpha=1$, $\gamma=0.2$, and (b) different $\gamma$ values with $\alpha=1$, $\beta=1$.
    Embeddings are derived from $\widehat{\textbf{h}}{}^i_{t-1}$.
    $Q{}^t_\text{trust}$ and $Q{}^t_\text{cont}$ across various $\beta$  (a) are comparable. 
    However, in (b), $Q{}^t_\text{cont}$ and $Q{}^t_\text{trust}$ show degradation, especially when $\gamma=0.3$. 
    Lower $\gamma$ values preserve neighborhoods similarly. 
    In Fig.~\ref{fig:rect_betgam_search}, both $\gamma=0.2$ and $\gamma=0.3$ led to good alignments; however, the neighborhood preservation deterioration with $\gamma=0.3$ prompts our choice of $\gamma=0.2$.}
    \label{fig:rect_quant_metrics}
\end{figure}

\begin{figure}
    \centering
    \includegraphics[width=0.95\linewidth]{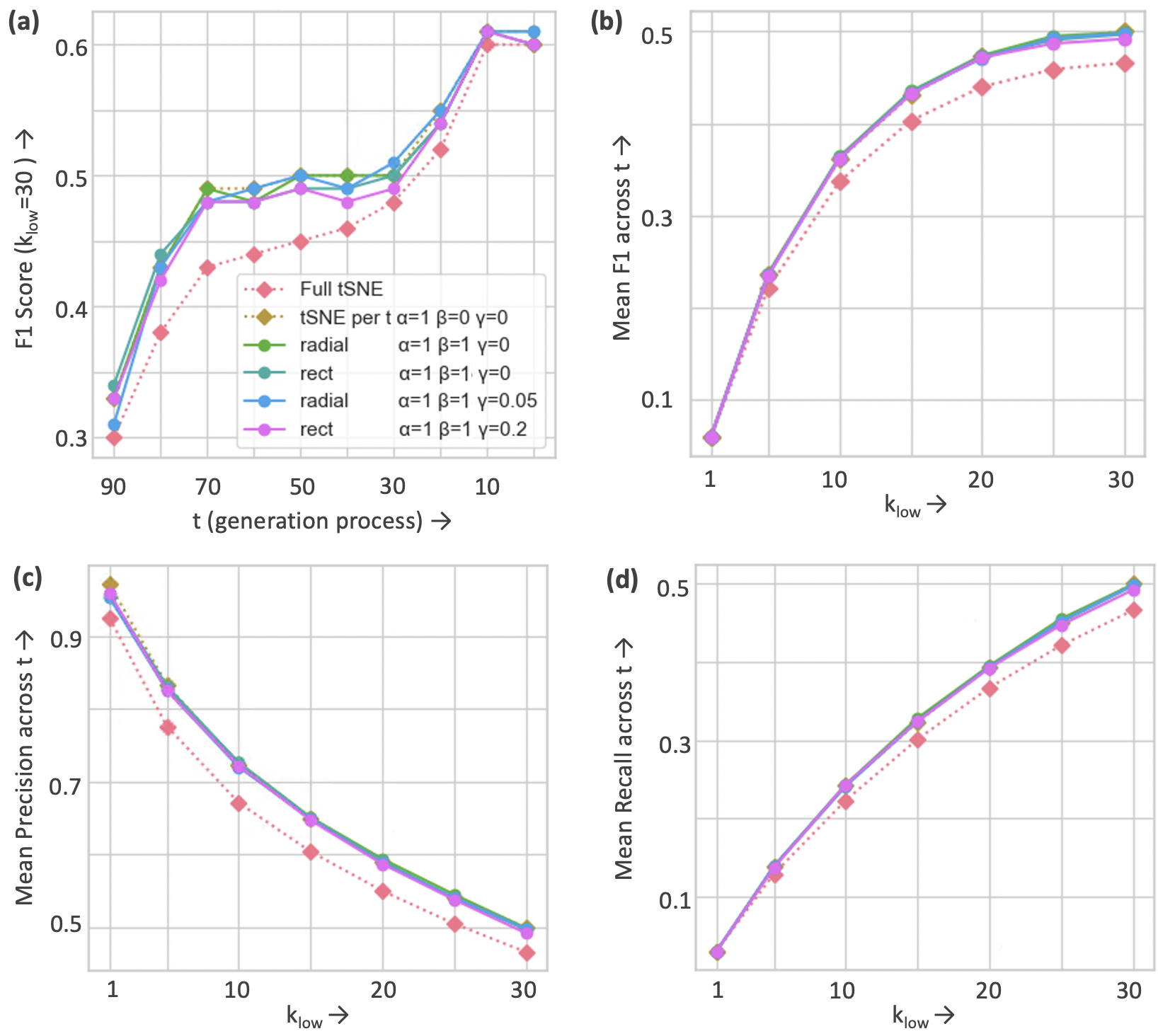}
    \caption{
    \textbf{Comparison of F1, Precision, and Recall for evolutionary embeddings vs. t-SNE baselines} across low-dimensional neighbours $k_{\text{low}}$,  $t$, and a fixed number of high-dimensional neighbours  $k_{\text{high}}$.
    (a) F1 score progression across time steps for $k_{\text{low}}=30$, comparing full vanilla t-SNE, stepwise vanilla t-SNE, and proposed radial and rectilinear layouts. (b-d) Average F1, Precision, and Recall across time steps for varying $k_{\text{low}}$. 
    The radial and rectilinear layouts with different $\beta$ and $\gamma$ values, perform similarly to stepwise t-SNE in preserving semantic relations in the low-dimensional space, while full vanilla t-SNE consistently underperforms across all metrics.}
    \label{fig:prec_recall_f1}
\end{figure}

\begin{figure}
    \centering
    \includegraphics[width=0.99\linewidth]{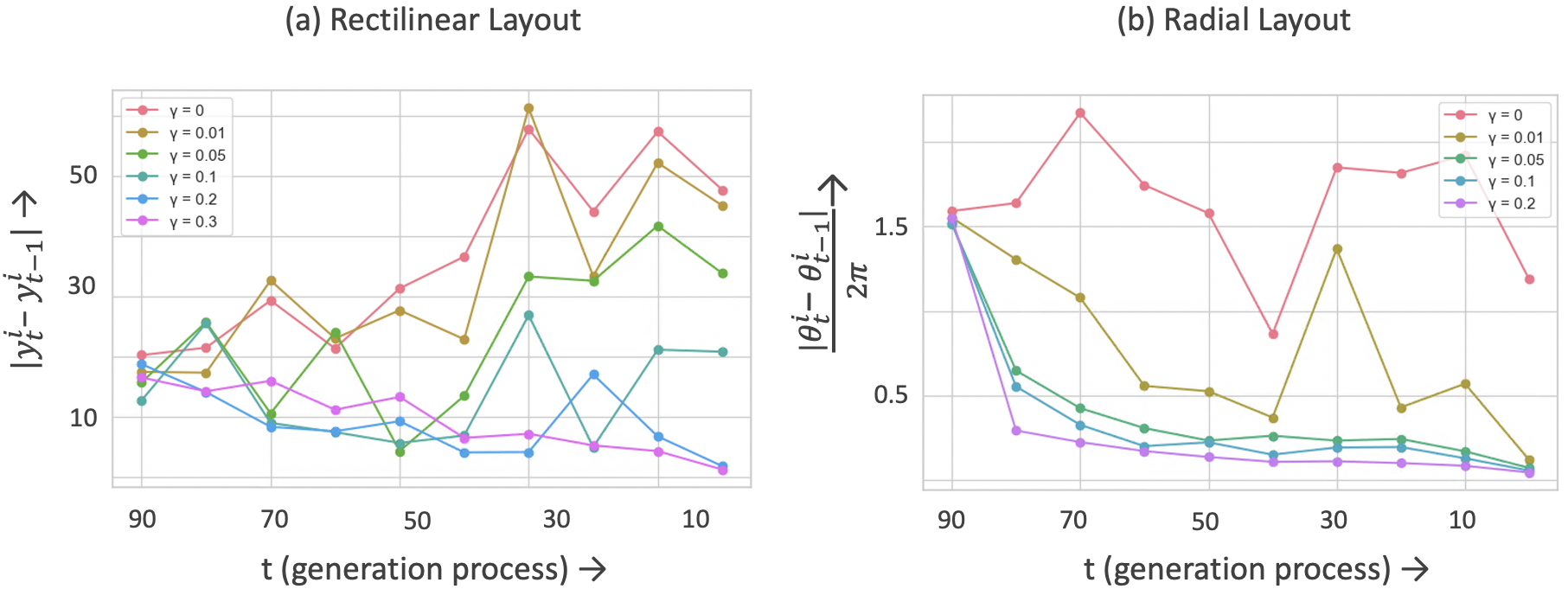}
    \caption{
    \textbf{Evaluation of the alignment constraint} on vertical and angular displacements across rectilinear (a) and radial (b) layouts across $\gamma$ values, with $\alpha=1$ and $\beta=1$.  Without the alignment, there is erratic behavior in $\gamma = 0$. We also see such behavior when the $\gamma$ values are small. 
    We select $\gamma = 0.2$ for rectilinear and $\gamma = 0.05$ for radial as the first stable values that do not degrade clustering quality seen in Fig.~\ref{fig:rect_quant_metrics} and~\ref{fig:rad_quant_metrics} respectively.
    The rectilinear layout shows more jumps compared to the radial layout, likely due to the radial layout's added flexibility for point movement in two directions around the ring. In contrast, the rectilinear layout limits movement inward, making optimization more challenging.
    Additionally, for stable $\gamma$ values ($\gamma<0.2$ for rectilinear and $\gamma<0.05$ for radial), alignment is poorer during the initial noisy steps (larger $t$) and gradually improves as generation progresses. This is likely due to the nature of the data itself, where initial steps generate more content with quickly changing clusters, while later steps focus on cluster refinement. }
    \label{fig:alignment}
\end{figure}

\begin{figure}
    \centering
    \includegraphics[width=0.95\linewidth]{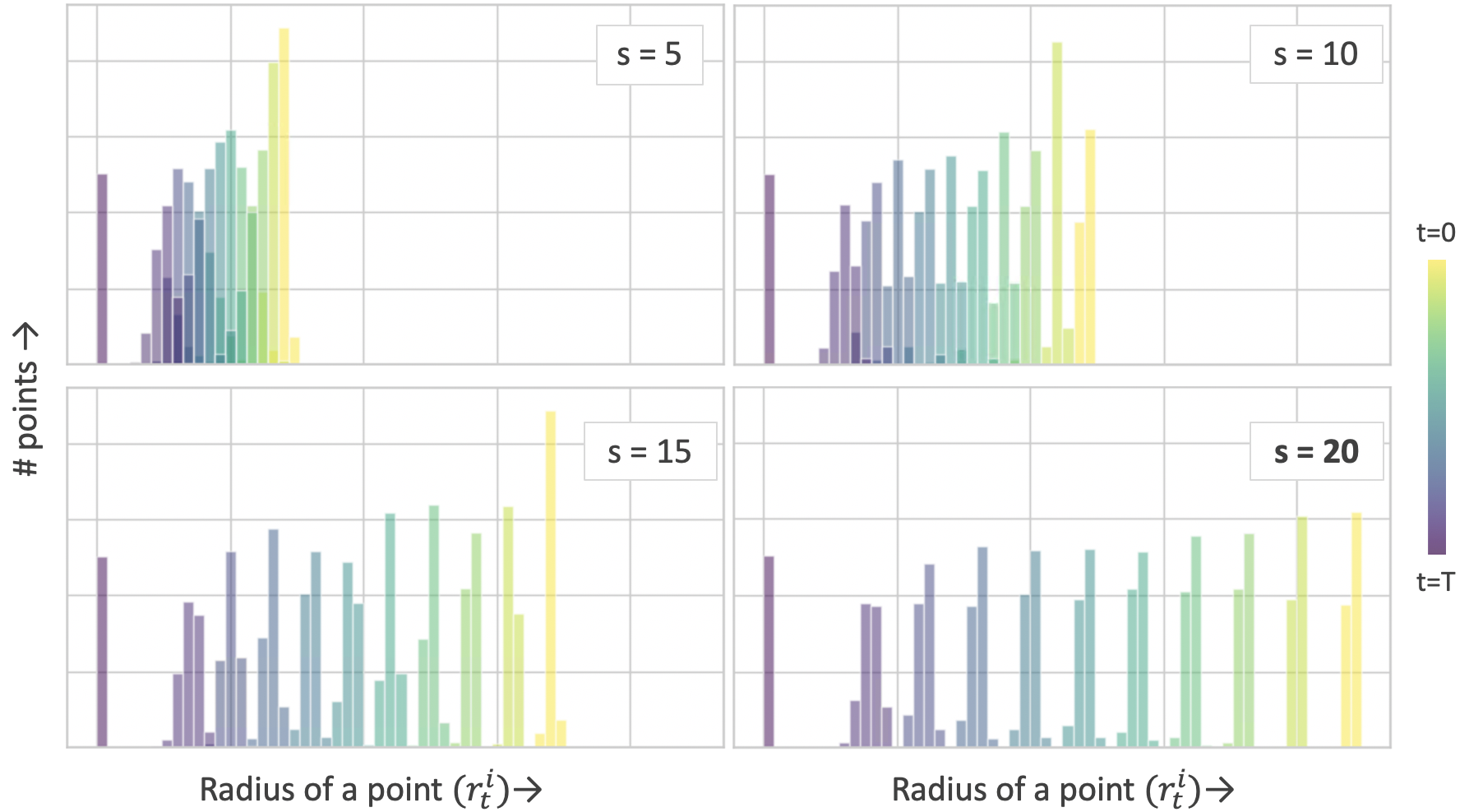}
    \caption{
    \textbf{Exploring the hyperparameter spacing $s$} used to set the ring offsets ($\bar{r}_t$) per iteration, $t$, where $\bar{r}_t=(T-t)*s$.
    Here, histograms illustrate the distribution of polar radii $r^i_t$ for radial embedding points $\textbf{l}^i_t$, projected from $\widehat{\textbf{h}}{}^i_t$ with $\alpha=1$, $\beta=1$, and $\gamma=0.05$. Each histogram is color-coded by its corresponding iteration $t$, with $T$ representing noise.
    A clear separation between iterations is observed when $s=20$, prompting us to use this value for experiments. 
    However, excessively large values of $s$ run the risk of insufficient allocation of space for early time steps.
    We observe similar insights for the rectilinear layout at $s=20$ hence we use the same value.}
    \label{fig:radius_acrossS}
\end{figure}

\begin{figure}
    \centering
    \includegraphics[width=0.95\linewidth]{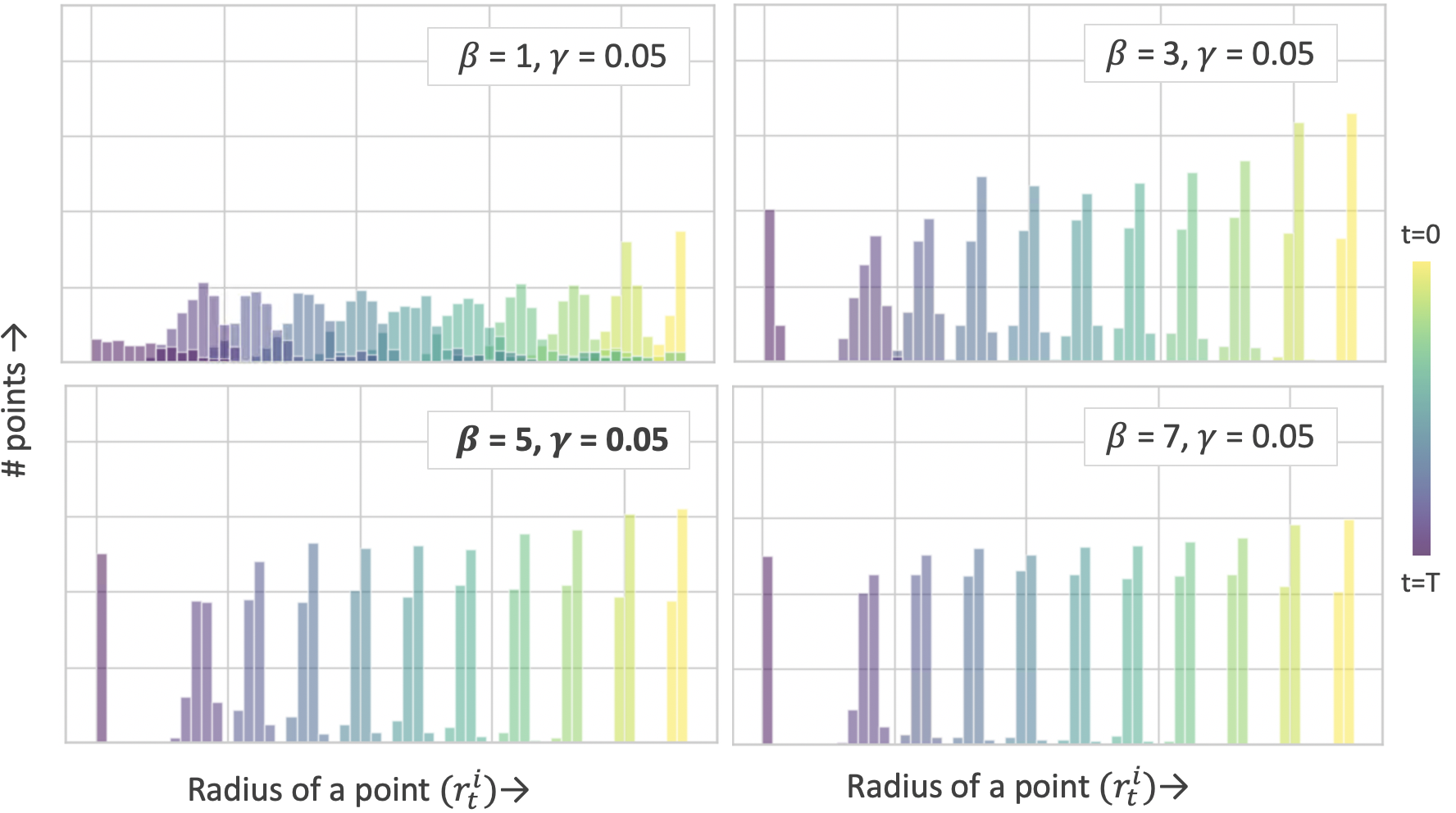}
    \caption{
    \textbf{Exploring the iteration separation across $\beta$ values} for the radial layout in a more systematic way. 
    Histograms are used to illustrate the distribution of polar radii $r^i_t$ for radial embedding points $\textbf{l}^i_t$, projected from $\widehat{\textbf{h}}{}^i_t$ with $\alpha=1$, $\gamma=0.05$, and $s=20$. 
    A clear separation between iterations is observed when $\beta=1$. 
    The same insights are used for the rectilinear layout at $\beta=1$.
    We use the same value for both the radial and rectilinear layouts.}
    \label{fig:radius_acrossB}
\end{figure}

\end{document}